\documentclass[11pt]{article}

\usepackage[preprint]{acl}

\usepackage{times}
\usepackage{latexsym}
\usepackage[hypcap=true]{caption}

\usepackage[T1]{fontenc}

\usepackage[utf8]{inputenc}

\usepackage{microtype}

\usepackage{inconsolata}

\usepackage{graphicx}
\usepackage{wrapfig}
\usepackage{amsmath}
\usepackage{amssymb}
\usepackage{multirow}
\usepackage{booktabs}
\usepackage{hyperref}
\usepackage{enumitem}
\usepackage{url}
\usepackage{subcaption} 
\usepackage{float}

\title{IRIS: Intent Resolution via Inference-time Saccades for Open-Ended VQA in Large Vision-Language Models}

\author{
Parsa Madinei $^{1,2}$\thanks{Equal contribution\\This work has been submitted to the IEEE for possible publication. Copyright may be transferred without notice, after which this version may no longer be accessible.} \qquad 
Srijita Karmakar $^{2}$\footnotemark[1] \\
\textbf{Russell Cohen Hoffing}$^{3}$ \qquad
\textbf{Felix Gervitz}$^{3}$ \qquad
\textbf{Miguel P. Eckstein}$^{1,2}$ \\
\\
$^{1}$ Department of Computer Science, UC Santa Barbara \\
$^{2}$ Department of Psychological \& Brain Sciences, UC Santa Barbara \\
$^{3}$ DEVCOM Army Research Laboratory \\
}

\begin{document}
\maketitle
\begin{abstract}
We introduce IRIS (Intent Resolution via Inference-time Saccades), a novel training-free approach that uses eye-tracking data in real-time to resolve ambiguity in open-ended VQA. Through a comprehensive user study with 500 unique image-question pairs, we demonstrate that fixations closest to the time participants start verbally asking their questions are the most informative for disambiguation in Large VLMs, more than doubling the accuracy of responses on ambiguous questions (from 35.2\% to 77.2\%) while maintaining performance on unambiguous queries. We evaluate our approach across state-of-the-art VLMs, showing consistent improvements when gaze data is incorporated in ambiguous image-question pairs, regardless of architectural differences. We release a new benchmark dataset to use eye movement data for disambiguated VQA, a novel real-time interactive protocol, and an evaluation suite.
\end{abstract}

\section{Introduction}

Visual question answering (VQA) represents a fundamental challenge at the intersection of Computer Vision and Natural Language Processing (NLP), requiring systems to understand both visual content and linguistic queries to generate appropriate responses. While recent Vision-Language Models (VLMs) have achieved impressive performance on standard VQA benchmarks \citep{chen2023pali, peng2023kosmos, chen2024spatialvlm, liu2024improved, wang2024cogvlm, qin2025efficient, yao2025efficient}, they continue to struggle with a pervasive real-world challenge: referential ambiguity. When multiple objects in an image could plausibly satisfy a query, such as asking "What is that?", current 
\begin{figure}
    \includegraphics[width=0.47\textwidth]{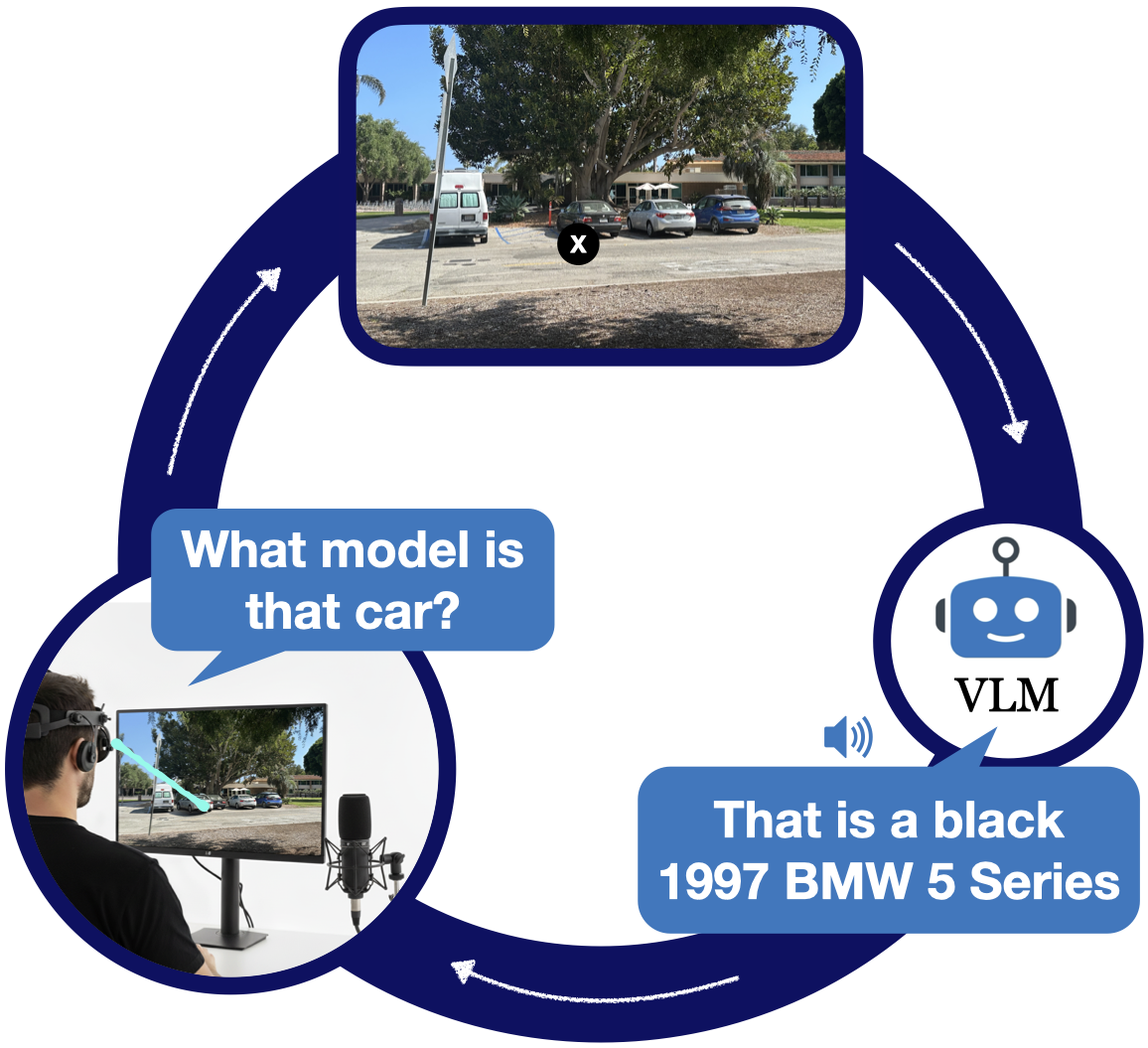}
    \caption{\textbf{IRIS overview}. Participant asks an ambiguous question about an image while their eyes are being tracked. The VLM uses the fixation data (marked as a white cross) to disambiguate the query and provide an accurate response in real-time. }
    \label{fig:system_overview}
    \vspace{5pt}
\end{figure}
VLMs lack the contextual grounding to identify the intended object referred to in the ambiguous query, defined as the "referent".


Here, we present a solution that leverages eye movement fixations, a natural human behavior, to resolve referential ambiguity in open-ended VQA. Decades of cognitive science and psycholinguistics research show a tight coupling between eye movements, attention allocation, and linguistic planning \citep{just1976eye,hayhoe2005eye,land2009vision, yarbus1967eye}. During natural viewing and questioning, fixations reliably precede verbal references by several hundred milliseconds, reflecting both planning and execution in speech production \citep{griffin2000eyes}. By aligning what is said with where (and when) people look, we obtain a time-locked, user-aligned signal that helps disambiguate referential intent in ambiguous VQA scenarios. 

\begin{figure*}[t]
  \includegraphics[width=1\textwidth]{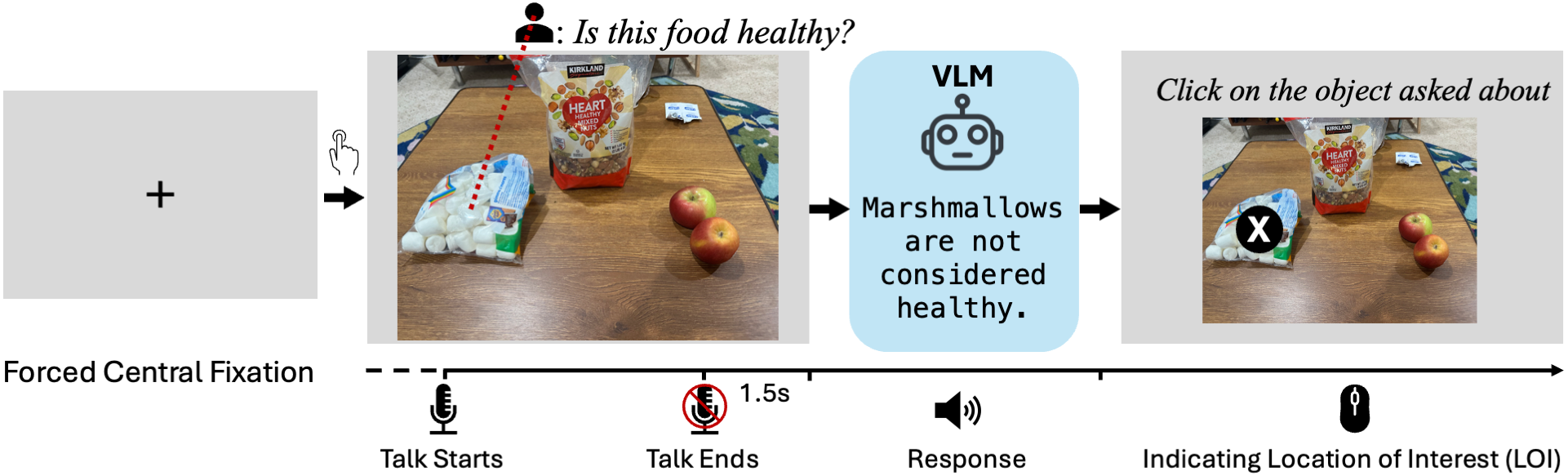}
  \caption{\textbf{Experimental procedure.} A central fixation check was enforced, after which participants freely viewed each image and asked any question aloud about it. Once 1.5s of silence elapsed following the question, the VLM was prompted with (i) the image, (ii) the transcribed question, and (iii) the same image with fixation data superimposed. Finally, participants reported the object they queried about (location of interest) by clicking the corresponding region of the image.}
  \label{fig:experiment_process}
\end{figure*}

We present IRIS, a training-free approach that leverages human gaze data to enhance VLMs' ability to resolve referential ambiguity in open-ended VQA. As illustrated in Figure~\ref{fig:system_overview}, our system captures users' natural eye movements as they formulate questions about images, then provides these fixation patterns as additional context to guide VLMs toward the intended referent.

Prior work leverages human gaze across vision–language and recognition tasks. For image captioning, \citet{sugano2016seeing} integrated human fixations into an attention LSTM; for VQA, \citet{inadumi2024gazevqa} estimated within-image gaze to select regions of interest, and \citet{sood2021gazevqa} assessed human–model attention alignment without feeding gaze to the model. Beyond VQA, \citet{karessli2017gazeembeddingszeroshotimage} encoded gaze as an auxiliary embedding to improve zero-shot image classification. In contrast, our approach operates at inference time without requiring model modification, making it immediately applicable to existing VLMs.

Through a comprehensive human study involving 500 unique image-question pairs, we collected a rich dataset of synchronized speech, gaze, and stimulus information. Participants formulated both ambiguous (for images containing multiple similar referents) and unambiguous questions (for images with a clear single referent), while we tracked their eye movements. Our experimental paradigm, detailed in Figure~\ref{fig:experiment_process}, captures the natural coupling between visual attention and linguistic formulation that occurs during human question-asking behavior.

Our temporal analysis reveals critical insights about the mechanism behind successful disambiguation using observers' gaze. We find that fixations occurring around the time of speech-onset provide the strongest disambiguation signals to VLMs. Moreover, we demonstrate that even a simple aggregation of all fixations during viewing significantly improves performance over image-only baselines ($p < .001$), suggesting that the concentration of gaze fixations itself carries disambiguating information. We evaluate our approach across 10 state-of-the-art (SOTA) VLMs, demonstrating consistent improvements when gaze data is incorporated. Our approach is not only grounded in literature linking eye movements and intent, but also practical and easily deployable. It finds immediate application in AR/VR systems that integrate eye tracking to deliver user-aligned ambiguity resolution in real-world interactive systems.

\noindent Our main contributions are:
\begin{itemize}[left=0pt, nosep]
    \item We introduce \textit{IRIS}, a training-free approach that uses human eye data to steer VLM representations toward user-intended referents, resolving ambiguity without modifying model parameters or reliance on a specific architecture.
    \item Across 500 human-collected image–question pairs, conditioning on gaze around speech onset \emph{more than doubles} accuracy for ambiguous questions (35.2\% $\rightarrow$ 77.2\%), while leaving unambiguous performance statistically unchanged. 
    \item A controlled analysis identifies the most informative window around speech onset, evidence that gaze aligns with task-relevant features when the ambiguous question is formulated.
    \item We create a new real-time interactive experimental paradigm that synchronizes speech, gaze, and image information for VQA, as well as a corresponding offline evaluation suite and a new dataset on open-ended VQA.
\end{itemize}

\begin{figure*}[t]
  \includegraphics[width=1\textwidth]{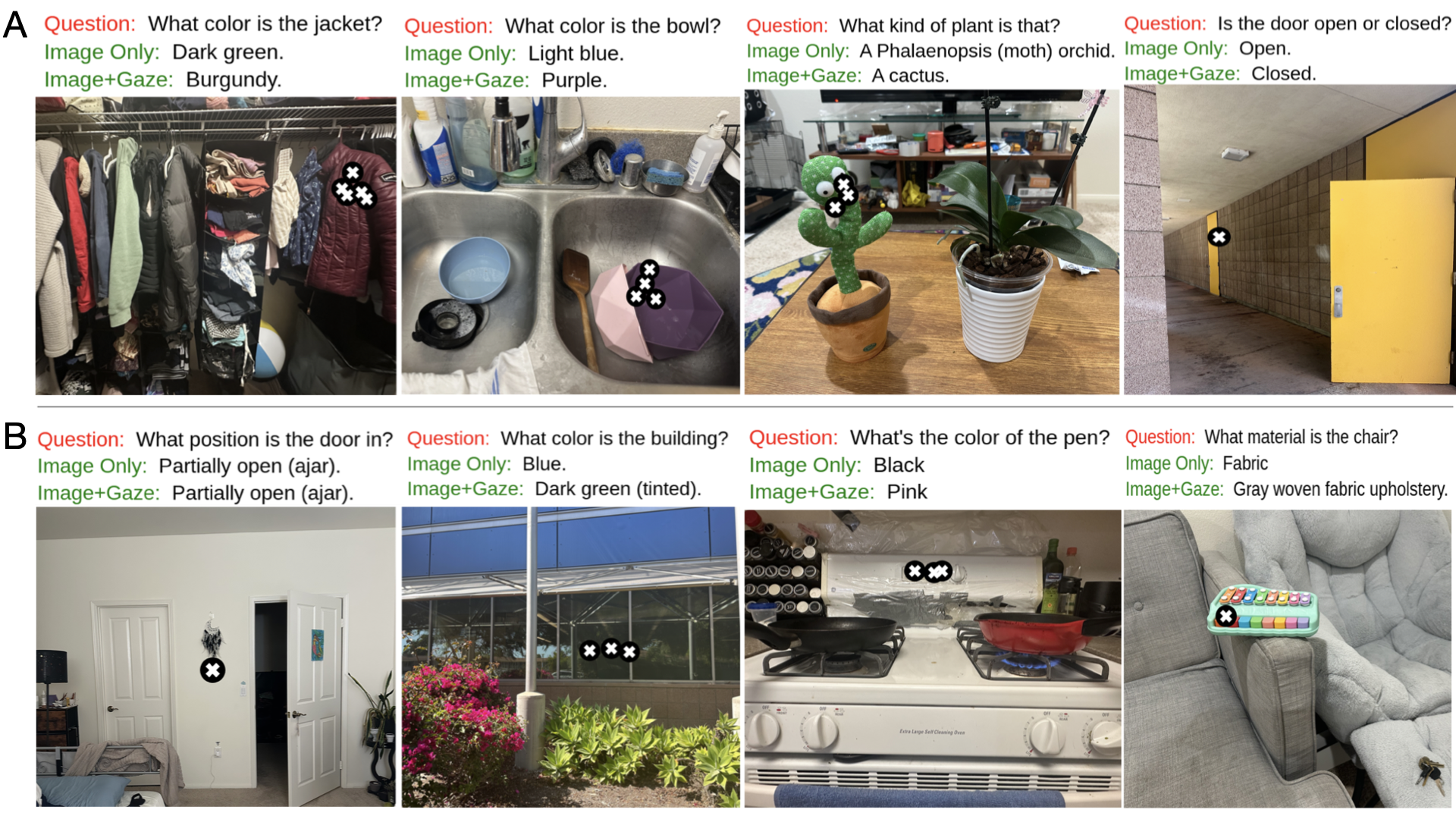}
  \caption{Qualitative results showing \textbf{successful (A) and failed (B) disambiguation using gaze data}. Black circles with white crosses mark temporally and spatially filtered fixation locations.}
  \label{fig:qualitative_results}
\end{figure*}

\section{IRIS}

\subsection{System Architecture}

Our IRIS system integrates three key components: (1) real-time eye-tracking to capture overt visual attention patterns (fixation locations), (2) speech recognition to identify question timing and content, and (3) MLLMs to generate responses. The system operates in a naturalistic setting where users freely explore images and formulate questions while their eye movements are recorded. The core insight driving our approach is that human gaze patterns during question formulation carry rich information about the intended referent of the question. When a person asks "What color is that?" while looking at a specific object among many, their fixations cluster around that object, providing a natural disambiguation signal. We capture these fixations and present them to the VLM as additional visual context, enabling the model to identify the referent and respond accurately.

\subsection{Gaze Data Collection Protocol}

We recruited 10 participants (5 female; age: 19-26 years, M=22±0.89 years) with normal or corrected-to-normal vision for a within-subjects study approved by the Institutional Review Board. All participants provided informed consent and received course credit for participation. Gaze was recorded monocularly from the left eye using EyeLink1000 Tower Mount \citep{SRResearch2005}, at 1000\,Hz (mean calibration error <1 dva; max <1.5 dva). Stimuli consisted of 50 photographs captured in everyday environments, with 40 scenes for ambiguous questions (containing multiple potential referents) and 10 for unambiguous ones (see Section~\ref{sec:appendixEx}, Figure~\ref{fig:figAex2} for examples). The images were blocked, and ambiguous and unambiguous blocks were counterbalanced across participants. We also included 7 practice images from the MS COCO test set \citep{lin2014mscoco}. Images were displayed on an Acer VG272X monitor ($1920\times1080$ @ 60 Hz; 60 cm width) at 62 cm viewing distance (0.029 dva/pixel). Each session began with a 9-point calibration procedure. Participants completed a practice block with feedback on valid questions. Trials were gated by a central forced-fixation check (Figure~\ref{fig:experiment_process}): participants pressed space while fixating on a cross; progression required fixating within a 1.45 dva radius, with recalibration after three failures. Once the fixation check was successful, participants were shown the trial image with no time limit while they formulated an unambiguous or ambiguous question (based on the block assigned). Participants then asked their question out loud while continuing to view the image. OpenAI's speech-to-text and text-to-speech APIs were used to convert participants' voice to text questions and VLM's text responses to voice output, respectively. Next, participants listened to the VLM's response. Finally, participants registered the Location of Interest (LOI), which was the location of the queried object using a mouse click. Sessions lasted 45 minutes per participant on average.

\subsection{Gaze Data Processing}

Gaze data typically consists of periods of relative stability, or fixations, which alternate with ballistic eye movements, also known as saccades. We identified fixations and saccades using EyeLink's online velocity and acceleration-based algorithm \citep{SRResearch2005} to parse saccades when eye velocity and acceleration exceed a threshold of 30°/$s$ and 8000°/$s^2$, respectively. 
The period between two saccades was classified as a fixation. The duration, start, and end locations of each fixation and saccade were recorded during the trial (see Section~\ref{sec:appendix2} and Figure~\ref{fig:figA2a} for descriptive statistics of recorded fixations). 

\begin{figure}[t]
  \includegraphics[width=0.5\textwidth]{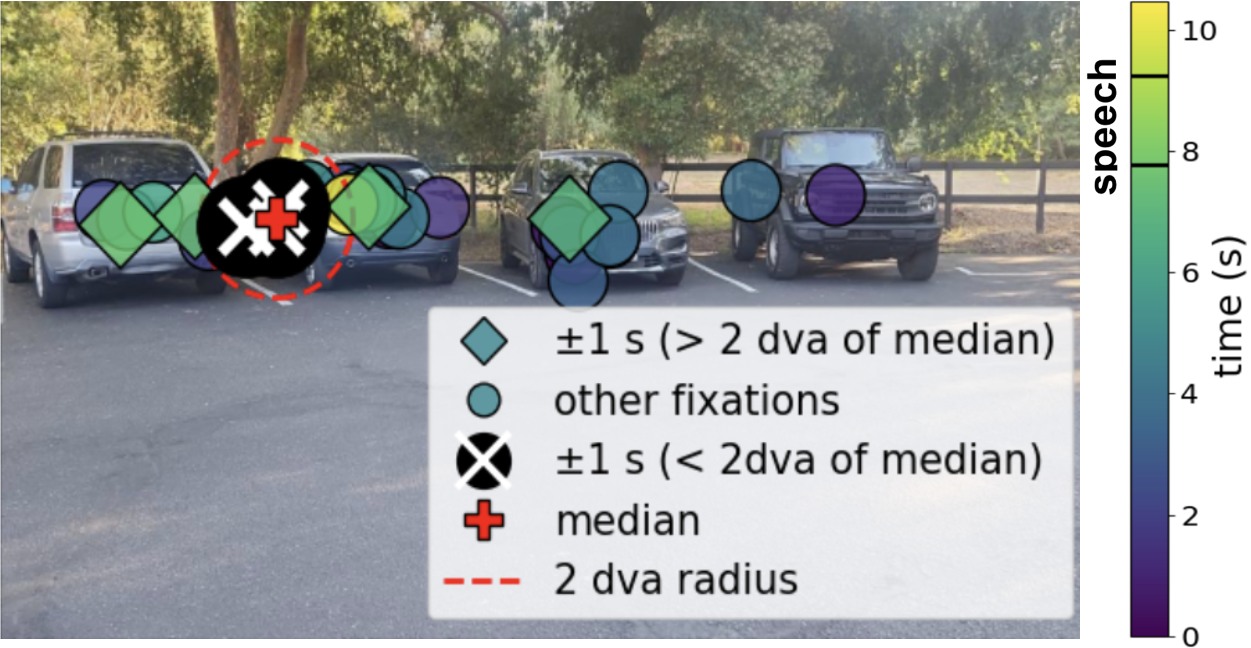}
  \caption{\textbf{Temporal-spatial filtering of eye gaze data.} Fixations colored by time. Black lines on the color bar mark speech onset and end. Diamonds represent fixations within ±1s of speech onset; all others are circles. Any diamond within 2 dva of the diamonds’ median (red +) is spatially filtered and rendered as white crosses on black circles and passed onto IRIS.}
  \label{fig:processing}
\end{figure}

For each trial, we used a spatiotemporal filtering approach to denoise fixation signals. First, we identified a specific time window around speech onset, which was determined using voice activity detection on the audio stream (see Section~\ref{sec:appendix2} and Figure~\ref{fig:figA2b} for details on speech onset latency and duration). Next, a spatial filter using the coordinate-wise median location of all temporally filtered fixations was applied. Any fixations that remained within a 2 dva radius of this median were kept, and the rest were discarded. If this step resulted in zero remaining fixations (10.8\% of all 500 trials), we kept all the fixations that passed the temporal filtering stage. Finally, we overlaid the filtered fixations on the original image as white crosses on black circles (Figure~\ref{fig:processing}). For more details on the efficiency of our filtering approach, see Section~\ref{sec:appendix1} (Figures~\ref{fig:figA1},  ~\ref{fig:figA1b}, and ~\ref{fig:figA1c}).

\subsection{VLM Integration}

We employed a system prompt that instructed the VLM to use eye movement data for disambiguation. Importantly, we instructed the model not to mention the eye movement data in its response, maintaining a natural interaction flow. The complete set of system prompts for all our tasks, including accuracy evaluation and baseline response generation (as described in Section~\ref{method:bounds}), is provided in \ref{sec:appendix0}.

\subsection{Ground Truth Generation}
\label{sec:groundtruth}
Establishing ground truth for open-ended VQA presents unique challenges, as responses can vary in phrasing while conveying the same information. We developed a multi-stage process for ground truth generation. We first generated responses from three SOTA VLMs, including GPT-5 \citep{openai2025gpt5}, Gemini 2.5 Pro \citep{gemini2025v25report}, and Claude Opus 4.1 \citep{anthropic2025opus41}, using the participants' LOI to identify the true object that was queried about. We then recruited five independent evaluators to review each image-question pair along with the model responses and the participants' LOI. They selected the most accurate response or provided a custom answer if none were satisfactory. This step minimized bias from any single model and incorporated human judgment for ambiguous cases. Because our semantic similarity metric (Section~\ref{sec:evaluation}) is sensitive to textual length and detail, we standardized the final ground truth by selecting the shortest accurate response among those chosen by evaluators. This approach ensured consistent comparisons across trials and conditions, while avoiding deflated similarity scores due to verbose phrasing. Finally, all ground truth responses were manually verified for accuracy by the first two authors.

\subsection{Evaluation Metrics} \label{sec:evaluation}

\begin{figure*}[t]
\centering
\includegraphics[width=1\textwidth]{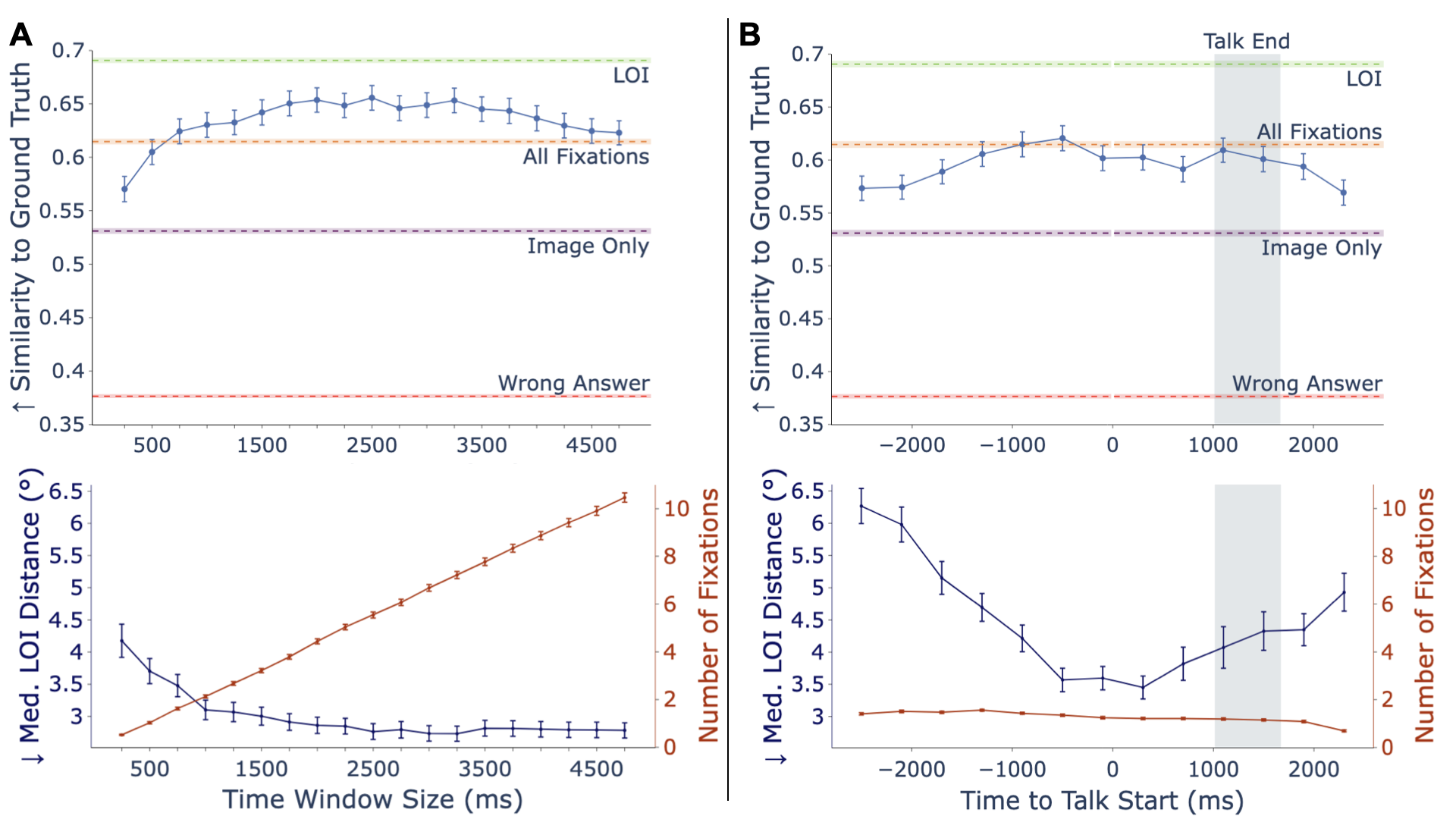}
\caption{\textbf{Temporal dynamics of gaze informativeness} for ambiguous questions. \textbf{(A) Model performance increases as the temporal window expands around speech onset time}, converging to the "all-fixations" baseline similarity at ±4500ms around speech onset. The bottom panel shows decreasing distance between the fixation median and the Location of Interest (LOI) in dark blue and increasing fixation count with larger windows in red. \textbf{(B) Peak performance a few milliseconds before speech onset} is revealed in sliding window analysis (600ms window-width, 400ms sliding step-size), aligning with minimum fixation-to-LOI distance. Gray shaded region indicates the interquartile range of speech end times. Error bars represent SEM. (A) and (B) top: green - "LOI (perfect gaze)" upper bound; orange - "all-fixations" baseline; purple - "image-only" baseline; pink - "wrong answer" lower bound (see Section~\ref{method:bounds} for details).}
\label{fig:temporal_analysis}
\end{figure*}

\begin{figure*}[t]
\centering
\includegraphics[width=\textwidth]{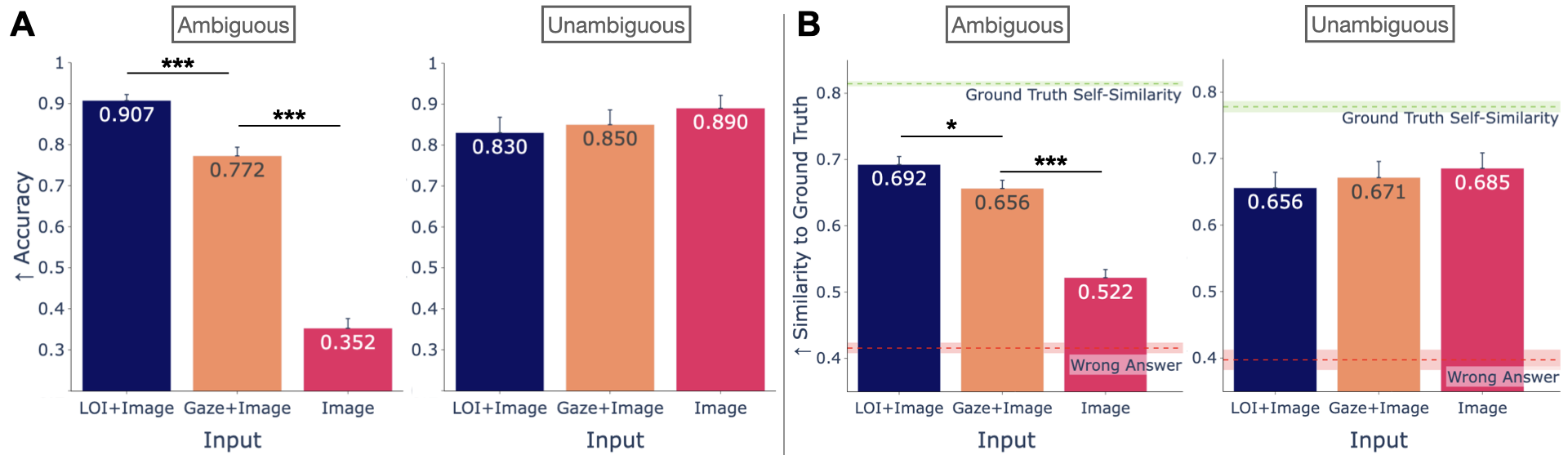}
\caption{\textbf{Effect of gaze input on performance} compared to Image-Only and Image with Location of Interest (LOI). (A) By adding gaze to the input, \textbf{accuracy increases from 35.2\% to 77.2\%, $p < .001$)} for ambiguous questions, while remaining non-significant for unambiguous questions ($p = .52$). (B) Semantic similarity shows similar significant improvements only in ambiguous conditions. Error bars represent SEM shown on the upper side.}
\label{fig:ambiguous_vs_unambiguous}
\end{figure*}

We employed two complementary metrics to assess system performance:

\textbf{Accuracy:} We evaluated response correctness using binary classification (correct vs.\ incorrect), assessed both manually and by an automated evaluator VLM, Gemini-2.5-Flash \citep{gemini2025flash}. The evaluator was provided with an accuracy evaluation prompt (see \ref{sec:appendix0}), alongside the question–response pair and the corresponding image annotated with the Location of Interest (LOI). Based on this input, it classified each response as correct or incorrect. To assess the reliability of this automated evaluation, we manually compared responses generated by GPT-5-Mini (results in Figure~\ref{fig:ambiguous_vs_unambiguous}) across all 500 image-question pairs 2and found an 88\% agreement in accuracy ratings between human judgment and the VLM evaluator. Unless noted, all $p$-values reported are from two-tailed paired-samples $t$ tests.

\textbf{Embedding-Based Similarity Metric:} Let $g_i$ denote the ground truth response for trial $i$ and $r_{i}^{(m)}$ the response of model $m$ on the same trial. We use a fixed, frozen sentence encoder $E(\cdot)$ to embed both texts \cite{openai2024embeddings}. Our primary score is cosine similarity outlined in \ref{eq:1}. For each model $m$, we report the mean and standard deviation of $s(\cdot,\cdot)$ across trials.
\begin{equation}
    s\!\left(r_{i}^{(m)}, g_i\right)
    \;=\;
    \frac{\langle E(r_{i}^{(m)}), E(g_i) \rangle}
         {\|E(r_{i}^{(m)})\|_2 \, \|E(g_i)\|_2}.
\label{eq:1}
\end{equation}

\subsection{Semantic Similarity Baselines, Upper and Lower Bounds}\label{method:bounds}

Semantic similarity values range between 0 (dissimilar) and 1 (identical) (Equation~\ref{eq:1}). To interpret the effects of gaze on ambiguous and unambiguous VQA, we compared the model's response to several baselines and upper and lower bounds. We generated baselines by sending specific information to the model (detailed here), and computing the similarity of the model response thus generated to the ground truth. Upper bounds: (i) mouse-click LOI as "perfect gaze" with image and question, and (ii) inter-rater agreement between the five human-generated ground truth statements (mean pairwise similarity). We also included a baseline that sends all fixations (no temporal or spatial filtering) to the model. Lower bounds: (i) only image and question only (no gaze) baseline and (ii) a "wrong-answer" baseline, generated by prompting the model to respond incorrectly given the image, LOI, and question (see Section~\ref{sec:appendix0}), to control for object-level differences within each image.

\section{Results}

\subsection{Qualitative Analysis}

Figure~\ref{fig:qualitative_results} qualitatively showcases when gaze helps and when it does not (see also Section~\ref{sec:appendix1}, Figures~\ref{fig:figA1} and ~\ref{fig:figA1b}). In Successful Cases (Figure~\ref{fig:qualitative_results}A), fixations narrow the model to the intended referent: e.g., gazing at the burgundy jacket corrects “Dark green,” fixating the purple bowl overrides a default to the salient blue one, fixations on the toy cactus disambiguate “plant,” and fixations on the closed door spatially resolve a two-door scene. In Failed Cases (Figure~\ref{fig:qualitative_results}B), gaze is uninformative or misleading: fixations fall between two doors, land on tinted glass yielding a wrong “Dark green” building color, reflect a voice-to-text error (“pen” vs “pan”) with vague gaze, or miss the relevant chair entirely. Overall, success requires fixations that are (1) concentrated on the target around speech onset, (2) show clear separation from distractors, and (3) aligned to question intent. Otherwise, gaze can have no effect, or even harm model performance. For a detailed taxonomy and quantification of errors, see Section~\ref{sec:appendix3} (Figure~\ref{fig:figA3a} shows examples of error types and Figure~\ref{fig:figA3b} quantifies that across all trials and participants).

\subsection{Temporal Analysis of Gaze Information}

To quantify features of gaze data that are: (i) crucial for successful disambiguation, and (ii) correlated with question intent, we systematically varied a temporal window centered on speech onset, to identify the optimal time window that contains the most informative gaze disambiguation signal (Figure~\ref{fig:temporal_analysis}). In Figure~\ref{fig:temporal_analysis}A, we observe that similarity to ground truth increases monotonically as we expand the temporal window from ±250ms to ±3000ms around speech onset. Performance starts at a similarity of 0.57 for the smallest window and reaches a peak of 0.65 at ±2500ms, after which it converges with the "all-fixations" baseline (0.615) at ±4500ms. This convergence suggests that fixations beyond ±4500ms around speech onset contribute minimal additional disambiguation value. The bottom panel reveals the mechanism underlying this performance curve: median fixation distance to LOI decreases from 4.0° to 2.7° as the window expands, while fixation count increases linearly. The inverse relationship between fixation-to-LOI distance and performance (correlation $r = -0.89, p < .001$) confirms that fixations closer to the intended object provide stronger disambiguation signals. Notably, even the all-fixations baseline similarity is significantly greater than that of image-only (0.615 vs 0.531, $p < .001$), indicating that the mere concentration of fixations during viewing helps resolve ambiguity compared to the absence of any gaze data. The LOI upper bound similarity (0.688) represents "perfect gaze" information, while the "wrong answer" baseline similarity (0.463) shows the lower bound by deliberately prompting the model to generate an incorrect response given image, LOI, and question (see Section~\ref{method:bounds}).

Figure~\ref{fig:temporal_analysis}B employs a sliding window approach (600ms width) to pinpoint when gaze is most informative in relation to speech onset. Performance peaks at a similarity of 0.62 for windows centered near speech onset (-200ms to +400ms to speech onset), precisely when the median fixation-to-LOI distance is minimal (3.5°). This temporal alignment between optimal performance and closest fixations to the intended object validates that speakers naturally look at what they are asking about. The performance decline and median fixation-to-LOI distance increase after speech onset suggest that later fixations introduce noise rather than a disambiguation signal. Manual accuracy analysis further confirmed that our temporal filtering approach yields a 17.7\% accuracy gain over the "all-fixations" baseline.

\subsection{Impact of Gaze on Ambiguous vs Unambiguous Questions}

We use the optimal time window identified in our analysis in Figure~\ref{fig:temporal_analysis} to demonstrate the differential impact of gaze information on ambiguous and unambiguous questions (Figure~\ref{fig:ambiguous_vs_unambiguous}). Gaze data within a 2-second window centered on speech onset provide substantial benefits when referential ambiguity exists, but offer minimal value when questions have clear, unique referents. For ambiguous questions (Figure~\ref{fig:ambiguous_vs_unambiguous}A, left), accuracy increases dramatically from 35.2\% (image-only) to 77.2\% (gaze+image), representing a 115\% improvement ($p < .001$). The large gap between image-only and gaze-augmented performance confirms that VLMs struggle with referential ambiguity but can effectively utilize gaze disambiguation signals when provided. In contrast, unambiguous questions show no significant improvement with gaze data (83.0\% to 86.0\%, $p = .52$). The minimal difference between conditions (LOI: 89.0\%, Gaze: 86.0\%, Image: 83.0\%) suggests a ceiling effect where the questions are sufficiently clear that additional context provides marginal benefit, as expected for unambiguous scenarios.

The semantic similarity analysis (Figure~\ref{fig:ambiguous_vs_unambiguous}B) corroborates these findings. Ambiguous questions show significant improvement from 0.531 (image) to 0.650 (gaze+image, $p < .001$), approaching the LOI performance of 0.691 ($p = .01$). The ground truth self-similarity of 0.82 represents the upper bound of achievable similarity given variation in human-generated ground truth responses. For unambiguous questions, similarity scores remain statistically unchanged across conditions (image: 0.656, gaze+image: 0.671, $p = .685$), reinforcing that gaze primarily benefits ambiguous scenarios. Importantly, the similarity-to-ground-truth measures for all conditions were greater than the "wrong answer" baseline (see Section~\ref{method:bounds}), indicating that deliberately guiding the model to answer incorrectly given the LOI leads to worse model performance than the image-only condition.

\begin{figure}[t]
  \includegraphics[width=0.5\textwidth]{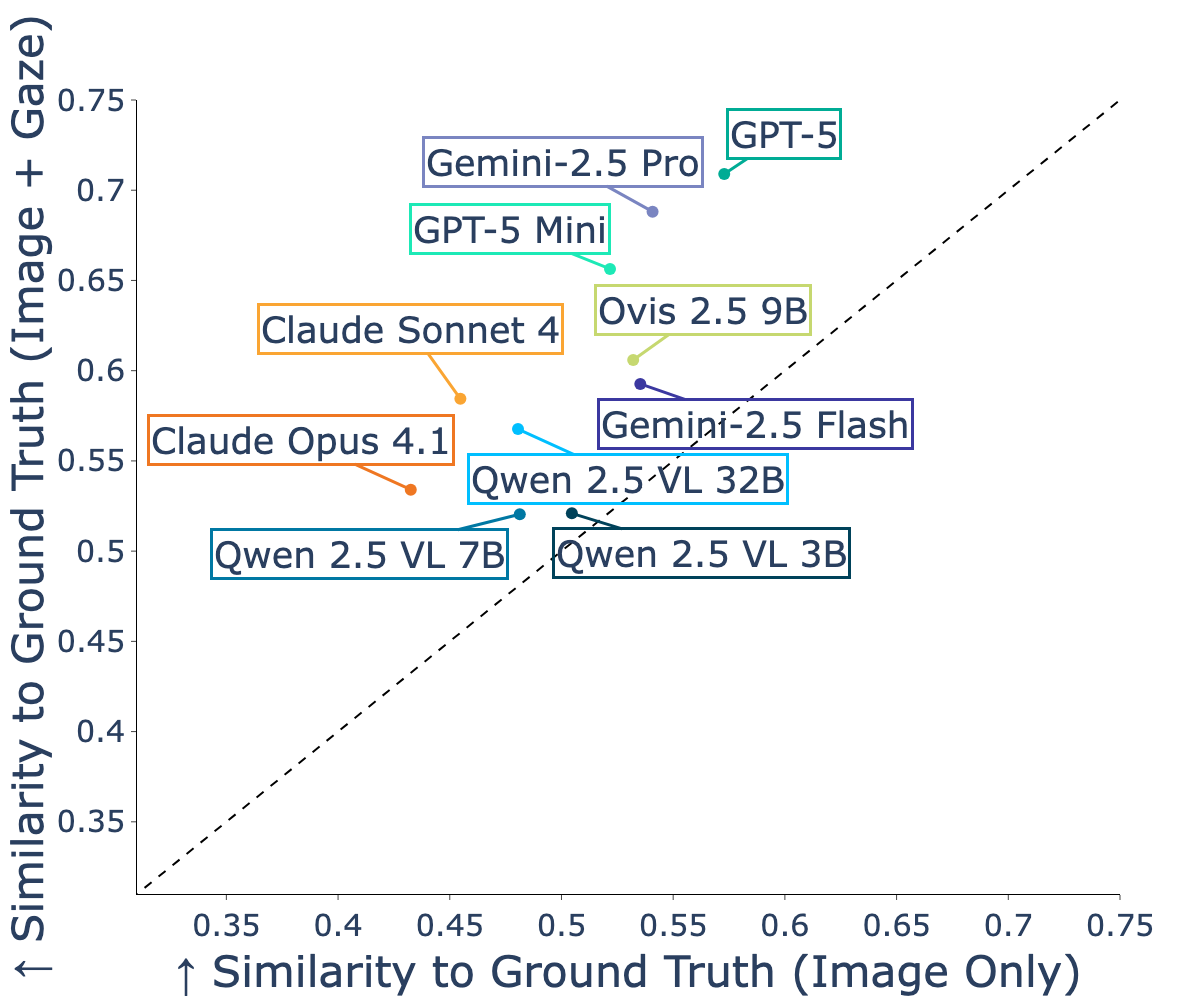}
  \caption{\textbf{Semantic similarity to ground truth across VLMs}. Models above the diagonal line show improvement with the inclusion of eye gaze data in ambiguous trials.}
  \label{fig:model_comparison}
\end{figure}

\subsection{Performance Across VLM Architectures}

Table~\ref{tab:ambig-acc} shows consistent accuracy gains on incorporating gaze data across diverse VLMs \citep{openai2025gpt5, gemini2025v25report, anthropic2025opus41, ovis25techreport, qwen25vl}. 
\begin{table*}[t]
\centering
\small
\setlength{\tabcolsep}{19pt}
\begin{tabular}{lcccc}
\toprule
\textbf{Model} & \textbf{Image Only (\%)} & \textbf{Image+Gaze (\%)} & \textbf{$\Delta$ (pp)} & \textbf{$p$-value} \\
\midrule
GPT-5 Mini            & 49.7  & \textbf{83.0}   & +33.2    & $<$\,0.001 \\
Gemini 2.5 Flash      & 54.2   & \textbf{83.3}   & +29.0    & $<$\,0.001 \\
Gemini 2.5 Pro        & 59.3   & \textbf{82.3}   & +23.0    & $<$\,0.001 \\
GPT-5                 & 53.7   & \textbf{76.5}   & +22.8    & $<$\,0.001 \\
Claude Sonnet 4       & 54.2  & \textbf{74.0}   & +19.8    & $<$\,0.001 \\
Ovis 2.5 9B           & 52.5   & \textbf{69.5}   & +17.0    & $<$\,0.001 \\
Claude 4.1 Opus       & 55.2  & \textbf{72.0}   & +16.7    & $<$\,0.001 \\
Qwen 2.5 VL 32B        & 56.2   & \textbf{73.0}   & +16.7    & $<$\,0.001 \\
Qwen 2.5 VL 7B        & 51.7   & \textbf{63.7}   & +12.0    & $<$\,0.001 \\
Qwen 2.5 VL 3B        & 52.0   & \textbf{54.5}   & +2.5    & 0.48 \\
\bottomrule
\end{tabular}
\caption{\textbf{VLM Accuracies on ambiguous trials.} ``Image+Gaze'' uses fixations from ±1s around speech onset; $\Delta$ is the difference in percentage points. $p$-values test \textit{Image+Gaze} vs.\ \textit{Image-Only}.}
\label{tab:ambig-acc}
\end{table*}

Across diverse architectures and model sizes, \emph{Image+Gaze} uniformly outperforms \emph{Image-only}, with improvements that are largest for frontier models and remain substantial for compact open-source ones. The effect correlates with instruction-following rather than parameter count alone, suggesting that gaze acts as a robust, architecture-agnostic disambiguation prior.
Figure~\ref{fig:model_comparison} shows the results of semantic similarity evaluation on the same models. Each point represents a model's performance with image-only similarity (x-axis) versus image+gaze similarity (y-axis) on ambiguous image-question pairs. Remarkably, all models show improvement with gaze data, falling above the diagonal identity line. GPT-5 achieves the highest absolute performance, reaching 0.71 similarity when augmented with gaze. These consistent across-family gains without fine-tuning or architectural changes suggest our method taps general visual-processing abilities and makes our approach immediately applicable to future VLMs as they continue to improve.

\subsection{Ablation Studies}

We conducted ablation studies to evaluate alternative methods for representing eye movement data to the VLM. Using GPT-5-Mini, we compared our baseline approach (white crosses overlaid on the image) against four alternative gaze representation methods (Table~\ref{tab:ablation}). \textbf{Heatmaps:} We generated Gaussian heatmaps from the eye movement data within the optimal time window and overlaid them on the original image with transparency. While heatmaps provide a continuous density representation of gaze attention, the smoothing inherent in heatmap generation may dilute the precise spatial information that discrete fixation markers preserve.
\textbf{Bounding Boxes:} We clustered the filtered fixations using DBSCAN and drew bounding boxes around the resulting clusters on the image. \textbf{Raw Coordinates as Text:} Instead of visual overlays, we provided the raw $(x, y)$ pixel coordinates of filtered fixations as text in the prompt alongside the unmodified image. \textbf{Cropped Images:} We cropped the image to a region centered on the median fixation location, discarding peripheral content. This approach performed the worst among all methods, as cropping removes potentially important contextual information and may eliminate parts of the queried object itself when fixations are offset from the object center.

\begin{table}[t]
\centering
\small
\setlength{\tabcolsep}{21pt}
\begin{tabular}{lc}
\toprule
\textbf{Method} & \textbf{Similarity} \\
\midrule
Crosses (Ours) & \textbf{0.830 $\pm$ 0.020} \\
Heatmap & 0.820 $\pm$ 0.019 \\
Bounding Box & 0.780 $\pm$ 0.021 \\
Coordinates (Text) & 0.748 $\pm$ 0.022 \\
Cropped & 0.703 $\pm$ 0.023 \\
\bottomrule
\end{tabular}
\caption{\textbf{Ablation study on gaze representation methods.} Semantic similarity to ground truth using GPT-5-Mini on ambiguous trials. Our cross-marker approach outperforms alternative representations.}
\label{tab:ablation}
\vspace{-10pt}
\end{table}

\section{Conclusion}


We present IRIS, a training-free approach that leverages human eye-tracking data to resolve referential ambiguity in vision-language models. Across 10 participants and 50 images, fixation data from a critical temporal window around speech onset more than doubled accuracy on ambiguous questions while maintaining performance on unambiguous ones. The consistent effectiveness across 10 diverse large VLMs demonstrates the generality and immediate applicability of our method. As eye-tracking technology becomes ubiquitous through AR/VR devices and accessibility tools, gaze-augmented VQA represents a promising paradigm for intuitive human-AI communication. By releasing our benchmark dataset, real-time experiment code, and evaluation suite, we provide the research community with tools to further explore gaze-informed VQA. Future work should explore extending this paradigm to other modalities and tasks, potentially revolutionizing how we interact with intelligent systems.

\section*{Limitations}
Our work represents an initial proof-of-concept for gaze-augmented VQA, and we intentionally scoped our study to controlled conditions to rigorously establish the core finding that natural gaze carries disambiguating information for VLMs. While we used research-grade eye-tracking equipment, the rapid advancement of consumer eye-tracking in devices suggests that the hardware requirements for IRIS-style systems are becoming increasingly accessible. Our participant sample, though sufficient to demonstrate statistically significant effects, reflects a university population; future work could expand demographic coverage to further validate generalizability. 

\bibliography{custom}

@misc{openai2025gpt5,
  title        = {Introducing GPT-5},
  author       = {OpenAI},
  year         = {2025},
  month        = aug,
  howpublished = {\url{https://openai.com/index/introducing-gpt-5/}},
  note         = {Accessed 2025-09-22}
}

@techreport{gemini2025v25report,
  title        = {Gemini 2.5: Pushing the Frontier with Advanced Reasoning, Multimodality, Long Context, and Next Generation Agentic Capabilities},
  author       = {{Gemini Team, Google DeepMind}},
  institution  = {Google DeepMind},
  year         = {2025},
  month        = jun,
  url          = {https://storage.googleapis.com/deepmind-media/gemini/gemini_v2_5_report.pdf},
  note         = {Covers Gemini 2.5 Pro and 2.5 Flash; Accessed 2025-09-22}
}

@misc{anthropic2025opus41,
  title        = {Claude Opus 4.1},
  author       = {Anthropic},
  year         = {2025},
  month        = aug,
  howpublished = {\url{https://www.anthropic.com/news/claude-opus-4-1}},
  note         = {Accessed 2025-09-22}
}

@article{ovis25techreport,
  title        = {Ovis2.5 Technical Report},
  author       = {Lu, S. and others},
  journal      = {arXiv preprint arXiv:2508.11737},
  year         = {2025},
  url          = {https://arxiv.org/abs/2508.11737}
}

@article{qwen25vl,
  title        = {Qwen2.5-VL Technical Report},
  author       = {Bai, S. and others},
  journal      = {arXiv preprint arXiv:2502.13923},
  year         = {2025},
  url          = {https://arxiv.org/abs/2502.13923},
  note         = {Single reference for Qwen2.5-VL 3B/7B/32B variants}
}

@misc{openai2024embeddings,
  author       = {OpenAI},
  title        = {OpenAI Text Embeddings},
  year         = {2024},
  howpublished = {\url{https://platform.openai.com/docs/guides/embeddings}},
  note         = {Accessed: September 2025}
}

@inproceedings{shanmuga2015eye,
  title={Eye tracking assisted extraction of attentionally important objects from videos},
  author={Shanmuga Vadivel, Karthikeyan and Ngo, Thuyen and Eckstein, Miguel and Manjunath, BS},
  booktitle={Proceedings of the IEEE Conference on Computer Vision and Pattern Recognition},
  pages={3241--3250},
  year={2015}
}

@article{reich2024role,
  title={{On the role of visual grounding in VQA}},
  author={Reich, Daniel and Schultz, Tanja},
  journal={arXiv preprint arXiv:2406.18253},
  year={2024}
}

@article{li2019visualbert,
  author    = {Li, Liunian Harold and Yatskar, Mark and Yin, Da and Hsieh, Cho{-}Jui and Chang, Kai{-}Wei},
  title     = {VisualBERT: A Simple and Performant Baseline for Vision and Language},
  journal   = {arXiv preprint arXiv:1908.03557},
  year      = {2019}
}

@inproceedings{tan2019lxmert,
  title = "{LXMERT}: Learning Cross-Modality Encoder Representations from Transformers",
    author = "Tan, Hao  and
      Bansal, Mohit",
    editor = "Inui, Kentaro  and
      Jiang, Jing  and
      Ng, Vincent  and
      Wan, Xiaojun",
    booktitle = "Proceedings of the 2019 Conference on Empirical Methods in Natural Language Processing and the 9th International Joint Conference on Natural Language Processing (EMNLP-IJCNLP)",
    month = nov,
    year = "2019",
    address = "Hong Kong, China",
    publisher = "Association for Computational Linguistics",
    url = "https://aclanthology.org/D19-1514/",
    doi = "10.18653/v1/D19-1514",
    pages = "5100--5111",
}

@inproceedings{radford2021clip,
  author    = {Radford, Alec and Kim, Jong Wook and Hallacy, Chris and Ramesh, Aditya and Goh, Gabriel and Agarwal, Sandhini and Sastry, Girish and Askell, Amanda and Mishkin, Pamela and Clark, Jack and Krueger, Gretchen and Sutskever, Ilya},
  title     = {Learning Transferable Visual Models From Natural Language Supervision},
  booktitle = {International Conference on Machine Learning},
  year      = {2021}
}

@inproceedings{alayrac2022flamingo,
  author    = {Alayrac, Jean{-}Baptiste and Donahue, Jeff and Luc, Pauline and Miech, Antoine and Barr, Iain and Hasson, Yana and Lenc, Karel and Mensch, Arthur and Millican, Katie and Reynolds, Malcolm and Ring, Roman and Rutherford, Eliza and Cabi, Serkan and Han, Tengda and Gong, Zhitao and Samangooei, Sina and Monteiro, Marianne and Menick, Jacob and Borgeaud, Sebastian and Brock, Andrew and Nematzadeh, Aida and Sharifzadeh, Sahand and Binkowski, Miko{\l}aj and Barreira, Ricardo and Vinyals, Oriol and Zisserman, Andrew and Simonyan, Karen},
  title     = {Flamingo: A Visual Language Model for Few-Shot Learning},
  booktitle = {NeurIPS},
  year      = {2022}
}

@inproceedings{li2023blip2,
  title={BLIP-2: Bootstrapping Language-Image Pre-training with Frozen Image Encoders and Large Language Models},
  author={Junnan Li and Dongxu Li and Silvio Savarese and Steven C. H. Hoi},
  booktitle={International Conference on Machine Learning},
  year={2023},
  url={https://api.semanticscholar.org/CorpusID:256390509}
}

@article{liu2023llava,
  author    = {Liu, Haotian and Li, Chunyuan and Wu, Qingyang and Lee, Yong Jae},
  title     = {Visual Instruction Tuning},
  journal   = {arXiv preprint arXiv:2304.08485},
  year      = {2023}
}

@article{li2024llavaonevision,
  author    = {Li, Bo and Zhang, Yuanhan and Guo, Dong and Zhang, Renrui and Li, Feng and Zhang, Hao and Zhang, Kaichen and Zhang, Peiyuan and Li, Yanwei and Liu, Ziwei and Li, Chunyuan},
  title     = {LLaVA-OneVision: Easy Visual Task Transfer},
  journal   = {arXiv preprint arXiv:2408.03326},
  year      = {2024}
}

@misc{wang2024qwen2vl,
    title={Qwen2-VL: Enhancing Vision-Language Model's Perception of the World at Any Resolution}, 
    author={Peng Wang and Shuai Bai and Sinan Tan and Shijie Wang and Zhihao Fan and Jinze Bai and Keqin Chen and Xuejing Liu and Jialin Wang and Wenbin Ge and Yang Fan and Kai Dang and Mengfei Du and Xuancheng Ren and Rui Men and Dayiheng Liu and Chang Zhou and Jingren Zhou and Junyang Lin},
    year={2024},
    eprint={2409.12191},
    archivePrefix={arXiv},
    primaryClass={cs.CV},
    url={https://arxiv.org/abs/2409.12191}
}

@misc{bai2025qwen2_5_vl,
    title={Qwen2.5-VL Technical Report}, 
    author={Shuai Bai and Keqin Chen and Xuejing Liu and Jialin Wang and Wenbin Ge and Sibo Song and Kai Dang and Peng Wang and Shijie Wang and Jun Tang and Humen Zhong and Yuanzhi Zhu and Mingkun Yang and Zhaohai Li and Jianqiang Wan and Pengfei Wang and Wei Ding and Zheren Fu and Yiheng Xu and Jiabo Ye and Xi Zhang and Tianbao Xie and Zesen Cheng and Hang Zhang and Zhibo Yang and Haiyang Xu and Junyang Lin},
    year={2025},
    eprint={2502.13923},
    archivePrefix={arXiv},
    primaryClass={cs.CV},
    url={https://arxiv.org/abs/2502.13923}
}

@misc{chen2024internvl25,
      title={Expanding Performance Boundaries of Open-Source Multimodal Models with Model, Data, and Test-Time Scaling}, 
      author={Zhe Chen and Weiyun Wang and Yue Cao and Yangzhou Liu and Zhangwei Gao and Erfei Cui and Jinguo Zhu and Shenglong Ye and Hao Tian and Zhaoyang Liu and Lixin Gu and Xuehui Wang and Qingyun Li and Yimin Ren and Zixuan Chen and Jiapeng Luo and Jiahao Wang and Tan Jiang and Bo Wang and Conghui He and Botian Shi and Xingcheng Zhang and Han Lv and Yi Wang and Wenqi Shao and Pei Chu and Zhongying Tu and Tong He and Zhiyong Wu and Huipeng Deng and Jiaye Ge and Kai Chen and Kaipeng Zhang and Limin Wang and Min Dou and Lewei Lu and Xizhou Zhu and Tong Lu and Dahua Lin and Yu Qiao and Jifeng Dai and Wenhai Wang},
      year={2025},
      eprint={2412.05271},
      archivePrefix={arXiv},
      primaryClass={cs.CV},
      url={https://arxiv.org/abs/2412.05271}, 
}

@article{laurencon2024idefics2,
  author    = {Lauren{\c{c}}on, Hugo and Tronchon, L{\'e}o and Cord, Matthieu and Sanh, Victor},
  title     = {What Matters When Building Vision-Language Models?},
  journal   = {arXiv preprint arXiv:2405.02246},
  year      = {2024}
}

@misc{mckinzie2024mm1,
      title={{MM1}: Methods, Analysis \& Insights from Multimodal LLM Pre-training}, 
      author={Brandon McKinzie and Zhe Gan and Jean-Philippe Fauconnier and Sam Dodge and Bowen Zhang and Philipp Dufter and Dhruti Shah and Xianzhi Du and Futang Peng and Floris Weers and Anton Belyi and Haotian Zhang and Karanjeet Singh and Doug Kang and Ankur Jain and Hongyu Hè and Max Schwarzer and Tom Gunter and Xiang Kong and Aonan Zhang and Jianyu Wang and Chong Wang and Nan Du and Tao Lei and Sam Wiseman and Guoli Yin and Mark Lee and Zirui Wang and Ruoming Pang and Peter Grasch and Alexander Toshev and Yinfei Yang},
      year={2024},
      eprint={2403.09611},
      archivePrefix={arXiv},
      primaryClass={cs.CV},
      url={https://arxiv.org/abs/2403.09611}, 
}

@inproceedings{tong2024eyeswideshut,
  author    = {Tong, Shengbang and Liu, Zhuang and Zhai, Yuexiang and Ma, Yi and LeCun, Yann and Xie, Saining},
  title     = {{Eyes wide shut? Exploring the visual shortcomings of multimodal LLMs}},
  booktitle = {CVPR},
  pages     = {9568--9578},
  year      = {2024}
}

@misc{prasad2024rephraseaugmentreasonvisual,
      title={Rephrase, Augment, Reason: Visual Grounding of Questions for Vision-Language Models}, 
      author={Archiki Prasad and Elias Stengel-Eskin and Mohit Bansal},
      year={2024},
      eprint={2310.05861},
      archivePrefix={arXiv},
      primaryClass={cs.CL},
      url={https://arxiv.org/abs/2310.05861}, 
}

@inproceedings{inadumi2024gazevqa,
  title     = {A Gaze-grounded Visual Question Answering Dataset for Clarifying Ambiguous {J}apanese Questions},
  author    = "Shun Inadumi and
               Seiya Kawano and
               Akishige Yuguchi and
               Yasutomo Kawanishi and
               Koichiro Yoshino",
  booktitle = "Proceedings of the 2024 Joint International Conference on Computational Linguistics, Language Resources and Evaluation (LREC-COLING 2024)",
  pages     = {558--571},
  year      = "2024"
}

@inproceedings{chen2024spatialvlm,
  title={Spatialvlm: Endowing vision-language models with spatial reasoning capabilities},
  author={Chen, Boyuan and Xu, Zhuo and Kirmani, Sean and Ichter, Brain and Sadigh, Dorsa and Guibas, Leonidas and Xia, Fei},
  booktitle={Proceedings of the IEEE/CVF Conference on Computer Vision and Pattern Recognition},
  pages={14455--14465},
  year={2024}
}

@article{yao2025efficient,
  title={Efficient GPT-4V level multimodal large language model for deployment on edge devices},
  author={Yao, Yuan and Yu, Tianyu and Zhang, Ao and Wang, Chongyi and Cui, Junbo and Zhu, Hongji and Cai, Tianchi and Chen, Chi and Li, Haoyu and Zhao, Weilin and others},
  journal={Nature Communications},
  volume={16},
  number={1},
  pages={5509},
  year={2025},
  publisher={Nature Publishing Group UK London}
}

@article{chen2023pali,
  title={Pali-x: On scaling up a multilingual vision and language model},
  author={Chen, Xi and Djolonga, Josip and Padlewski, Piotr and Mustafa, Basil and Changpinyo, Soravit and Wu, Jialin and Ruiz, Carlos Riquelme and Goodman, Sebastian and Wang, Xiao and Tay, Yi and others},
  journal={arXiv preprint arXiv:2305.18565},
  year={2023}
}

@inproceedings{liu2024improved,
  title={Improved baselines with visual instruction tuning},
  author={Liu, Haotian and Li, Chunyuan and Li, Yuheng and Lee, Yong Jae},
  booktitle={Proceedings of the IEEE/CVF conference on computer vision and pattern recognition},
  pages={26296--26306},
  year={2024}
}

@article{peng2023kosmos,
  title={Kosmos-2: Grounding multimodal large language models to the world},
  author={Peng, Zhiliang and Wang, Wenhui and Dong, Li and Hao, Yaru and Huang, Shaohan and Ma, Shuming and Wei, Furu},
  journal={arXiv preprint arXiv:2306.14824},
  year={2023}
}

@article{wang2024cogvlm,
  title={Cogvlm: Visual expert for pretrained language models},
  author={Wang, Weihan and Lv, Qingsong and Yu, Wenmeng and Hong, Wenyi and Qi, Ji and Wang, Yan and Ji, Junhui and Yang, Zhuoyi and Zhao, Lei and XiXuan, Song and others},
  journal={Advances in Neural Information Processing Systems},
  volume={37},
  pages={121475--121499},
  year={2024}
}

@book{yarbus1967eye,
  author    = {Alfred L. Yarbus},
  title     = {Eye Movements and Vision},
  year      = {1967},
  publisher = {Springer},
  address   = {New York, NY},
  doi       = {10.1007/978-1-4899-5379-7},
  url       = {https://link.springer.com/book/10.1007/978-1-4899-5379-7},
  isbn      = {978-1-4899-5379-7},
  edition   = {1},
  note      = {SpringerLink eBook edition published 2013}
}

@article{just1976eye,
  title={Eye fixations and cognitive processes},
  author={Just, Marcel Adam and Carpenter, Patricia A},
  journal={Cognitive psychology},
  volume={8},
  number={4},
  pages={441--480},
  year={1976},
  publisher={Elsevier}
}

@article{hayhoe2005eye,
  title={Eye movements in natural behavior},
  author={Hayhoe, Mary and Ballard, Dana},
  journal={Trends in cognitive sciences},
  volume={9},
  number={4},
  pages={188--194},
  year={2005},
  publisher={Elsevier}
}

@article{land2009vision,
  title={Vision, eye movements, and natural behavior},
  author={Land, Michael F},
  journal={Visual neuroscience},
  volume={26},
  number={1},
  pages={51--62},
  year={2009},
  publisher={Cambridge University Press}
}

@article{griffin2000eyes,
  title={What the eyes say about speaking},
  author={Griffin, Zenzi M and Bock, Kathryn},
  journal={Psychological science},
  volume={11},
  number={4},
  pages={274--279},
  year={2000},
  publisher={SAGE Publications Sage CA: Los Angeles, CA}
}

@article{sugano2016seeing,
  title={Seeing with humans: Gaze-assisted neural image captioning},
  author={Sugano, Yusuke and Bulling, Andreas},
  journal={arXiv preprint arXiv:1608.05203},
  year={2016}
}

@techreport{gemini2025flash,
  author    = {Gemini Team, Google},
  title     = {Gemini 2.5: Pushing the Frontier with Advanced Reasoning, Multimodality, Long Context, and Next Generation Agentic Capabilities},
  institution = {Google DeepMind},
  year      = {2025},
  note      = {Gemini 2.5 Flash model},
  url       = {https://storage.googleapis.com/deepmind-media/gemini/gemini_v2_5_report.pdf}
}

@article{tanenhaus1995integration,
  title={Integration of visual and linguistic information in spoken language comprehension},
  author={Tanenhaus, Michael K and Spivey-Knowlton, Michael J and Eberhard, Kathleen M and Sedivy, Julie C},
  journal={Science},
  volume={268},
  number={5217},
  pages={1632--1634},
  year={1995}
}

@article{borji2013state,
  title={State-of-the-art in visual attention modeling},
  author={Borji, Ali and Itti, Laurent},
  journal={IEEE transactions on pattern analysis and machine intelligence},
  volume={35},
  number={1},
  pages={185--207},
  year={2013}
}

@article{itti2001computational,
  title={Computational modelling of visual attention},
  author={Itti, Laurent and Koch, Christof},
  journal={Nature reviews neuroscience},
  volume={2},
  number={3},
  pages={194--203},
  year={2001}
}

@article{jacob1991use,
author = {Jacob, Robert J. K.},
title = {The use of eye movements in human-computer interaction techniques: what you look at is what you get},
year = {1991},
issue_date = {April 1991},
publisher = {Association for Computing Machinery},
address = {New York, NY, USA},
volume = {9},
number = {2},
issn = {1046-8188},
url = {https://doi.org/10.1145/123078.128728},
doi = {10.1145/123078.128728},
journal = {ACM Trans. Inf. Syst.},
month = apr,
pages = {152–169},
numpages = {18}
}

@incollection{majaranta2014eye,
  author    = {P{\"a}ivi Majaranta and Andreas Bulling},
  title     = {Eye Tracking and Eye-Based Human--Computer Interaction},
  booktitle = {Advances in Physiological Computing},
  editor    = {Stephen H. Fairclough and Kiel Gilleade},
  series    = {Human--Computer Interaction Series},
  publisher = {Springer},
  address   = {London},
  year      = {2014},
  doi       = {10.1007/978-1-4471-6392-3_3},
  url       = {https://doi.org/10.1007/978-1-4471-6392-3_3}
}

@misc{sukthanker2018anaphora,
      title={Anaphora and Coreference Resolution: A Review}, 
      author={Rhea Sukthanker and Soujanya Poria and Erik Cambria and Ramkumar Thirunavukarasu},
      year={2018},
      eprint={1805.11824},
      archivePrefix={arXiv},
      primaryClass={cs.CL},
      url={https://arxiv.org/abs/1805.11824}, 
}

@misc{mao2016generation,
      title={Generation and Comprehension of Unambiguous Object Descriptions}, 
      author={Junhua Mao and Jonathan Huang and Alexander Toshev and Oana Camburu and Alan Yuille and Kevin Murphy},
      year={2016},
      eprint={1511.02283},
      archivePrefix={arXiv},
      primaryClass={cs.CV},
      url={https://arxiv.org/abs/1511.02283}
}

@inproceedings{kazemzadeh2014referitgame,
    title = "{R}efer{I}t{G}ame: Referring to Objects in Photographs of Natural Scenes",
    author = "Kazemzadeh, Sahar  and
      Ordonez, Vicente  and
      Matten, Mark  and
      Berg, Tamara",
    editor = "Moschitti, Alessandro  and
      Pang, Bo  and
      Daelemans, Walter",
    booktitle = "Proceedings of the 2014 Conference on Empirical Methods in Natural Language Processing ({EMNLP})",
    month = oct,
    year = "2014",
    address = "Doha, Qatar",
    publisher = "Association for Computational Linguistics",
    url = "https://aclanthology.org/D14-1086/",
    doi = "10.3115/v1/D14-1086",
    pages = "787--798"
}

@misc{zhang2016yin,
      title={Yin and Yang: Balancing and Answering Binary Visual Questions}, 
      author={Peng Zhang and Yash Goyal and Douglas Summers-Stay and Dhruv Batra and Devi Parikh},
      year={2016},
      eprint={1511.05099},
      archivePrefix={arXiv},
      primaryClass={cs.CL},
      url={https://arxiv.org/abs/1511.05099}, 
}

@article{coco2012scan,
  title={Scan patterns predict sentence production in the cross-modal processing of visual scenes},
  author={Coco, Moreno I and Keller, Frank},
  journal={Cognitive science},
  volume={36},
  pages={1204--1223},
  year={2012}
}

@article{qin2025efficient,
  title={Efficient knowledge distillation and alignment for improved KB-VQA},
  author={Qin, Xiaofei and Pei, Ruiqi and He, Changxiang and Li, Fan and Zhang, Xuedian},
  journal={Scientific Reports},
  volume={15},
  number={1},
  pages={20682},
  year={2025},
  publisher={Nature Publishing Group UK London}
}

@article{hayhoe2012predictive,
  title={Predictive eye movements in natural vision},
  author={Hayhoe, Mary M and McKinney, Travis and Chajka, Kelly and Pelz, Jeff B},
  journal={Experimental brain research},
  volume={217},
  number={1},
  pages={125--136},
  year={2012},
  publisher={Springer}
}

@misc{testoni2024racquet,
      title={RACQUET: Unveiling the Dangers of Overlooked Referential Ambiguity in Visual LLMs}, 
      author={Alberto Testoni and Barbara Plank and Raquel Fernández},
      year={2024},
      eprint={2412.13835},
      archivePrefix={arXiv},
      primaryClass={cs.CL},
      url={https://arxiv.org/abs/2412.13835}, 
}

@manual{SRResearch2005,
  author       = {{SR Research Ltd.}},  
  year         = {2005},
  title        = {EyeLink 1000 [Apparatus and software]},
  organization = {{SR Research Ltd.}},
  url          = {https://www.sr-research.com/},
}

@article{posner1980orienting,
  author  = {Posner, Michael I.},
  title   = {Orienting of attention},
  journal = {Quarterly Journal of Experimental Psychology},
  year    = {1980},
  volume  = {32},
  number  = {1},
  pages   = {3--25},
  doi     = {10.1080/00335558008248231}
}

@article{hoffman1995role,
  author  = {Hoffman, James E. and Subramaniam, Babu},
  title   = {The role of visual attention in saccadic eye movements},
  journal = {Perception \& Psychophysics},
  year    = {1995},
  volume  = {57},
  number  = {6},
  pages   = {787--795},
  doi     = {10.3758/BF03206794}
}

@article{deubel1996saccade,
  author  = {Deubel, Heiner and Schneider, Werner X.},
  title   = {Saccade target selection and object recognition: Evidence for a common attentional mechanism},
  journal = {Vision Research},
  year    = {1996},
  volume  = {36},
  number  = {12},
  pages   = {1827--1837},
  doi     = {10.1016/0042-6989(95)00294-4}
}

@article{li2021dissociating,
  author  = {Li, Hsin-Hung and Hanning, Nina M. and Carrasco, Marisa},
  title   = {To look or not to look: dissociating presaccadic and covert spatial attention},
  journal = {Trends in Neurosciences},
  year    = {2021},
  volume  = {44},
  number  = {8},
  pages   = {669--686},
  doi     = {10.1016/j.tins.2021.05.002},
  pmid    = {34099240},
  pmcid   = {PMC8552810},
  url     = {https://pmc.ncbi.nlm.nih.gov/articles/PMC8552810/}
}

@inproceedings{lin2014mscoco,
  author    = {Lin, Tsung{-}Yi and Maire, Michael and Belongie, Serge and Hays, James and Perona, Pietro and Ramanan, Deva and  Zitnick, C. Lawrence and Doll{\'a}r, Piotr},
  title     = {Microsoft COCO: Common Objects in Context},
  booktitle = {Computer Vision -- ECCV 2014},
  year      = {2014},
  editor    = {Fleet, David and Pajdla, Tom{\'a}{\v{s}} and Schiele, Bernt and Tuytelaars, Tinne},
  series    = {Lecture Notes in Computer Science},
  volume    = {8693},
  pages     = {740--755},
  publisher = {Springer},
  address   = {Cham},
  doi       = {10.1007/978-3-319-10602-1_48}
}

@inproceedings{
ilaslan2023gazevqa,
title={Gaze{VQA}: A Video Question Answering Dataset for Multiview Eye-Gaze Task-Oriented Collaborations},
author={Muhammet Furkan Ilaslan and Chenan Song and Joya Chen and Difei Gao and Weixian Lei and Qianli Xu and Joo Hwee Lim and Mike Zheng Shou},
booktitle={The 2023 Conference on Empirical Methods in Natural Language Processing},
year={2023},
url={https://openreview.net/forum?id=MkD0VGShAq}
}

@article{sood2021gazevqa,
  author       = {Ekta Sood and
                  Fabian K{\"{o}}gel and
                  Florian Strohm and
                  Prajit Dhar and
                  Andreas Bulling},
  title        = {{VQA-MHUG:} {A} Gaze Dataset to Study Multimodal Neural Attention
                  in Visual Question Answering},
  journal      = {CoRR},
  volume       = {abs/2109.13116},
  year         = {2021},
  url          = {https://arxiv.org/abs/2109.13116},
  eprinttype    = {arXiv},
  eprint       = {2109.13116},
  bibsource    = {dblp computer science bibliography, https://dblp.org}
}

@misc{karessli2017gazeembeddingszeroshotimage,
      title={Gaze Embeddings for Zero-Shot Image Classification}, 
      author={Nour Karessli and Zeynep Akata and Bernt Schiele and Andreas Bulling},
      year={2017},
      eprint={1611.09309},
      archivePrefix={arXiv},
      primaryClass={cs.CV},
      url={https://arxiv.org/abs/1611.09309}, 
}

\pagebreak
\appendix
\section{Appendix}\label{sec:appendix}

\subsection{Examples of Ambiguous and Unambiguous Questions}\label{sec:appendixEx}

\begin{figure*}[t]
    \centering
    \includegraphics[width=\textwidth]{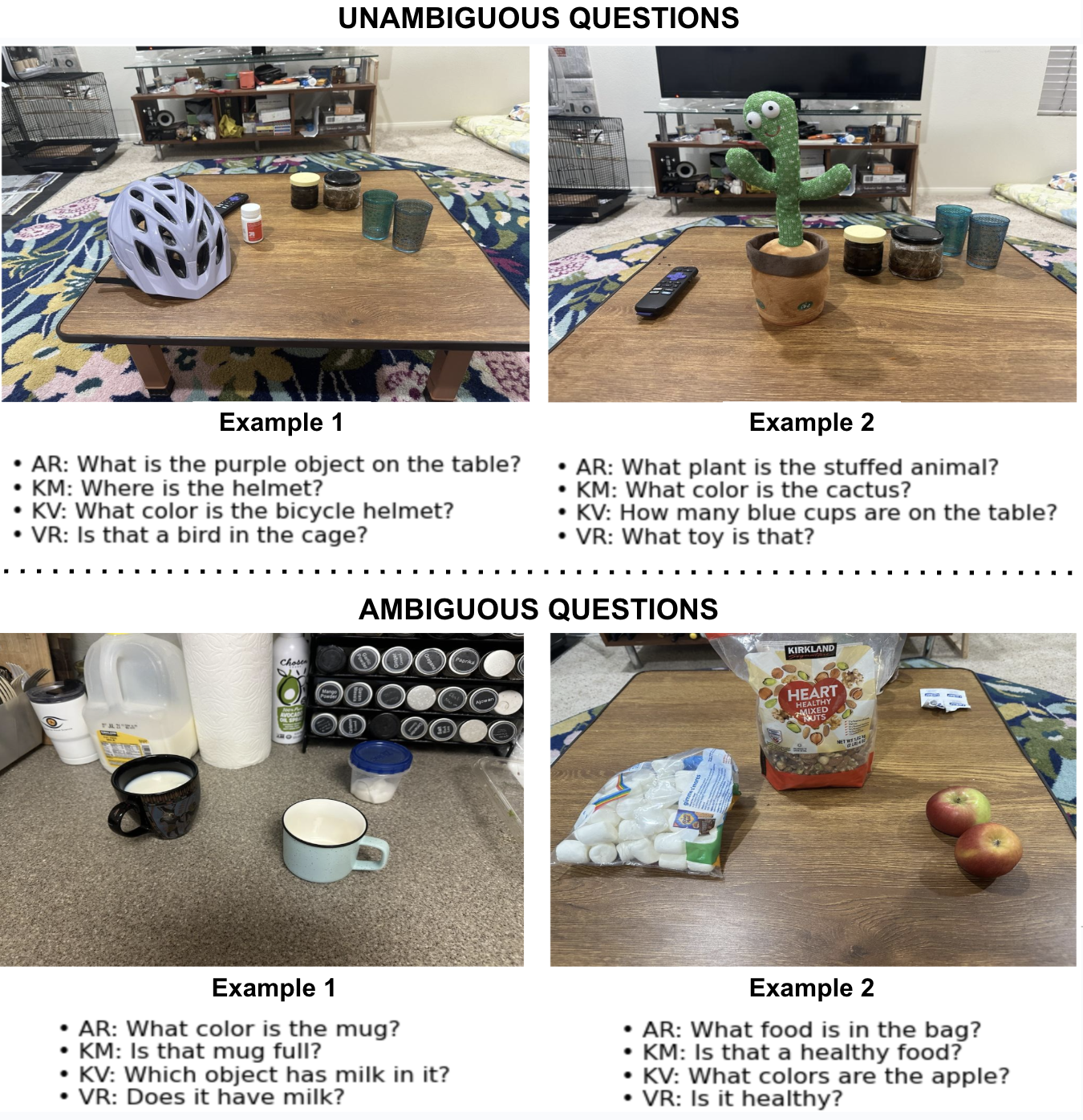}
    \caption{\textbf{Sample unambiguous and ambiguous questions} asked by four participants on four example images.}
    \label{fig:figAex2}
\end{figure*}

Each of our 10 participants formulated their own questions for the same 50 images (10 for unambiguous questions and 40 for ambiguous ones), thus resulting in 500 unique image-question pairs. Figure~\ref{fig:figAex2} shows sample questions on a few example images. 

The top panel shows two examples from the Unambiguous condition. For instance, in Unambiguous Example 1, participant AR asks about "the purple object," and there is only one such object present in the image. In Unambiguous Example 2, participant KM asks about the "cactus," again unambiguously referring to the only cactus in the image.

In contrast, the bottom panel shows examples of ambiguous questions. For instance, in Ambiguous Example 1, participants AR and KM both ask about a mug, but it is unclear from their questions which one of the two mugs present in the image is being referred to, making their questions ambiguous given the image information. Similarly, in Ambiguous Example 2, participants KM and VR inquire whether the "food" is healthy, but there are several food items present, of which some are healthy (apples) but others are not (marshmallows), thus making their question ambiguous about the intended referent. KV's question on this image is not ambiguous and, thus, inappropriate for our study (see Section~\ref{sec:appendix3})

\subsection{Temporo-Spatial Visualization of Raw Fixation Data}\label{sec:appendix1}

\begin{figure*}[t]
    \centering
    \includegraphics[width=\textwidth]{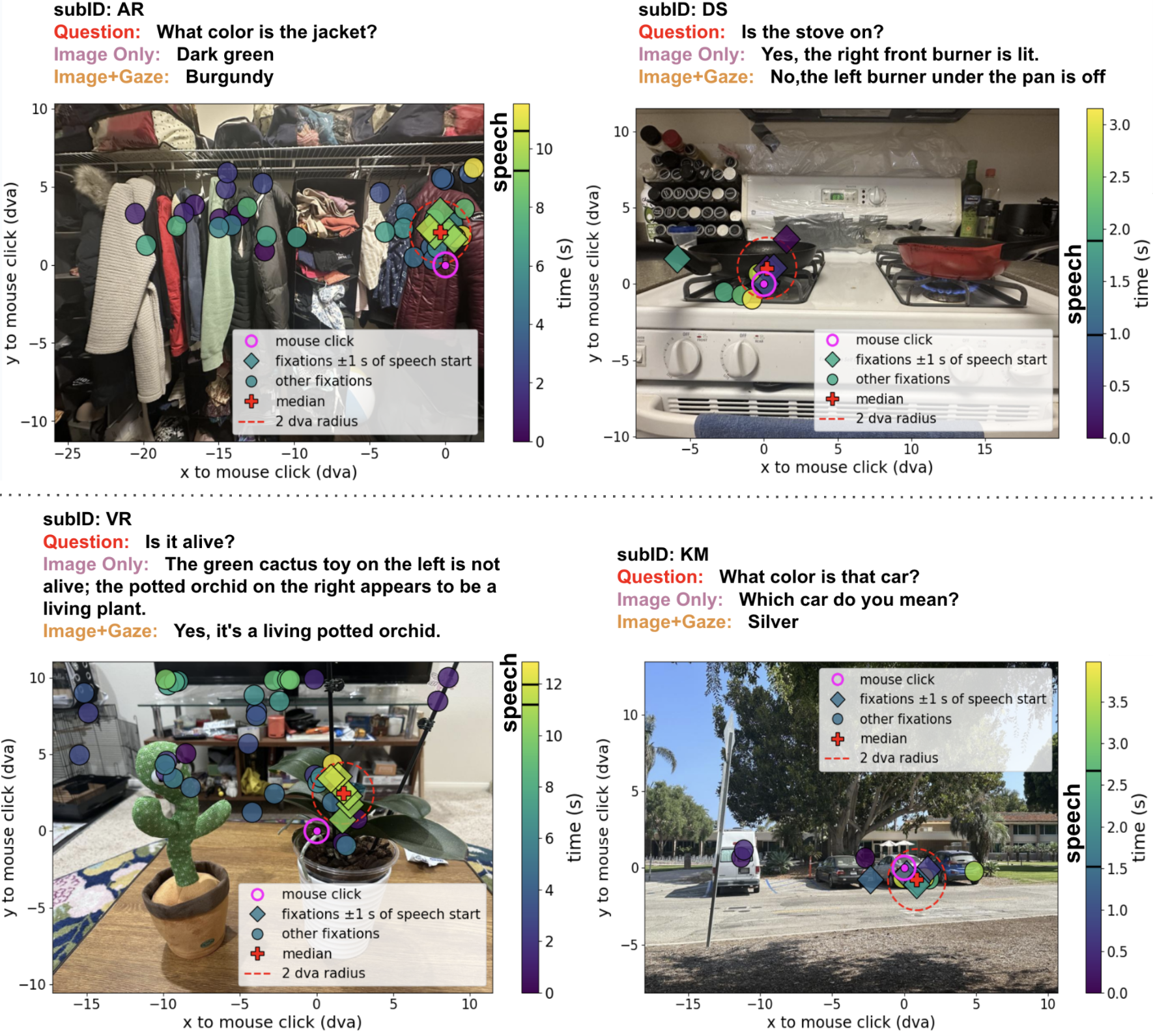}
    \caption{\textbf{Visualizing and filtering fixations: examples of "successful" ambiguous trials.} Inclusion of gaze data successfully disambiguated the object in question. The colored circles and diamonds are fixations shaded by their temporal order (yellow - more recent, blue - older). The magenta circle represents the mouse click location registered by the participant at the end of the trial. The red dashed circle indicates a region of 2 dva radius around the median of all fixations within ±1s around speech onset (shown as diamonds). Any diamond within the red dashed circle was sent to the VLM as eye gaze data. The x and y-axes indicate distance to the mouse click location. The times at which the participant started and ended their question are marked as black lines on the color bar. The fixation markers are enlarged for illustrative purposes.}
    \label{fig:figA1}
\end{figure*}

\begin{figure*}[t]
    \centering
    \includegraphics[width=0.9\textwidth]{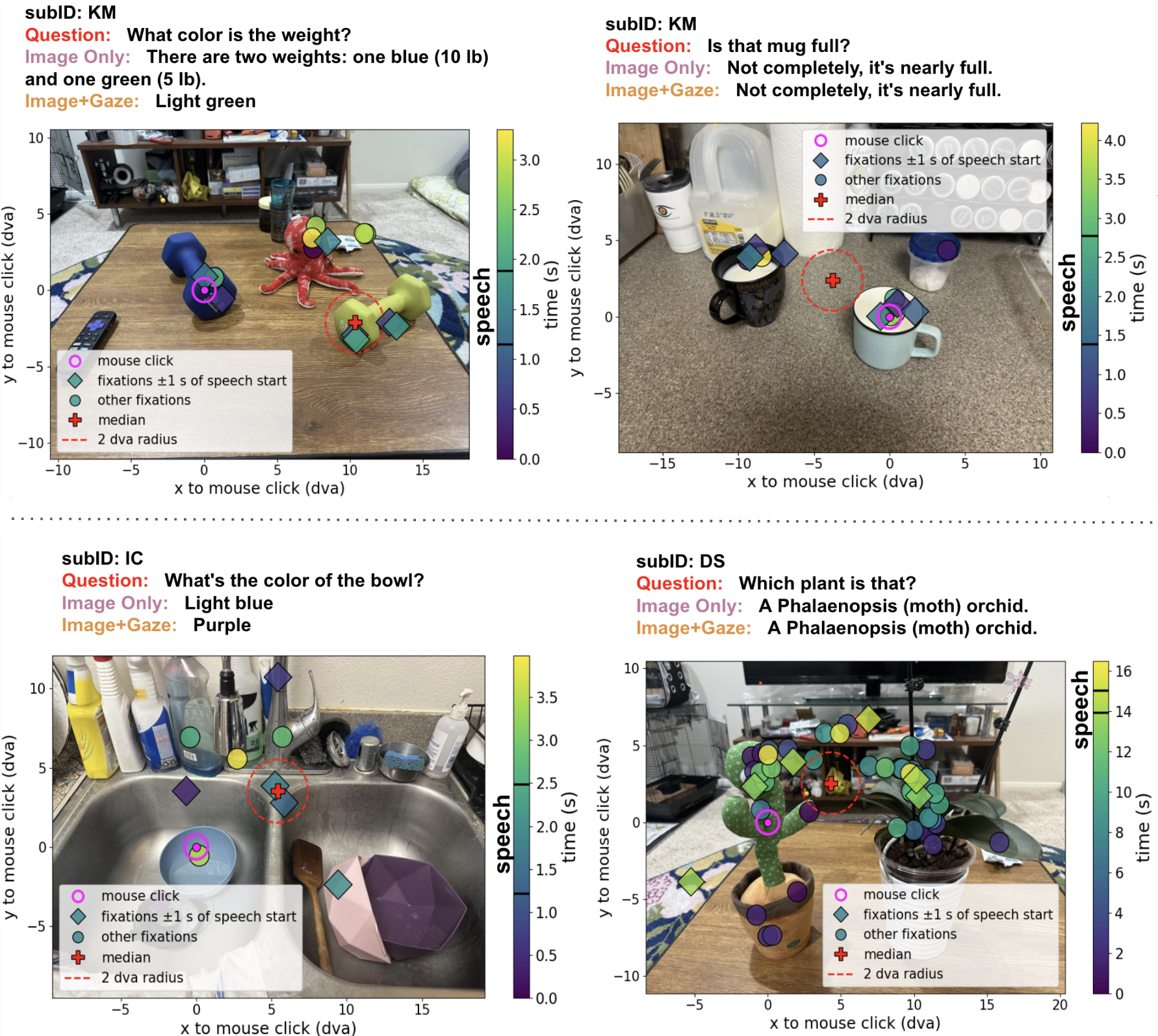}
    \caption{\textbf{Visualizing and filtering fixations: examples of "unsuccessful" ambiguous trials.} Inclusion of eye gaze data did not successfully disambiguate the object in these trials. Plotting conventions follow those described in Figure~\ref{fig:figA1}.}
    \label{fig:figA1b}
\end{figure*}

\begin{figure*}[t]
    \centering
    \includegraphics[width=0.9\textwidth]{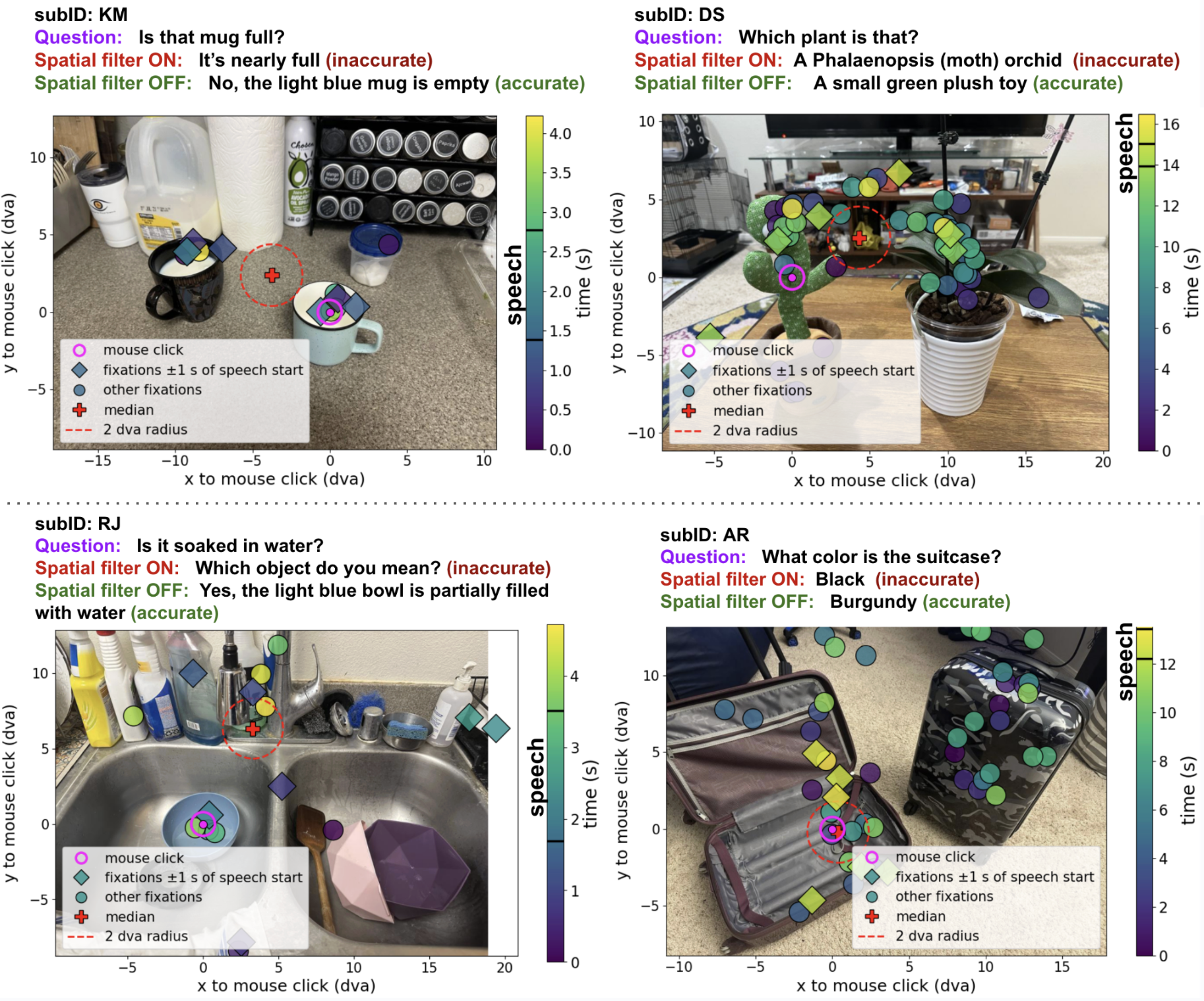}
    \caption{\textbf{Effect of spatial filtering.} Visualizing ambiguous Image+Gaze trials where spatial filtering resulted in zero retained fixations. In these examples, removing the spatial filter (but retaining the temporal filter) led to accurate disambiguation (2\% trials). In other words, all fixations marked as diamonds were sent to the VLM during inference. Plotting conventions follow those described in Figure~\ref{fig:figA1}.}
    \label{fig:figA1c}
\end{figure*}

We plot the raw fixation start locations on trial images to visualize the temporal and spatial features of our eye movement data with respect to task-related events such as speech onset time and mouse click location. Figure~\ref{fig:figA1} shows example successful trials, where the eye gaze data helped the VLM disambiguate the object being asked about in ambiguous scenarios. Figure~\ref{fig:figA1b} shows example unsuccessful trials, where the inclusion of eye gaze data did not disambiguate the object in question for our VLM and algorithm. Finally, Figure~\ref{fig:figA1c} describes examples of a few rare cases (1\% of all trials) where our spatial filtering approach resulted in discarding all fixations and lead to an incorrect VLM response, but keeping all fixations that occurred within ±1s of speech onset rescued the trial. Importantly, the data shown are from different participants, emphasizing the generality of our approach to eye gaze data and questions collected across subjects.

In the figure above, we show all fixations made by a participant during the trial as circle and diamond markers, colored according to the temporal order of their occurrence. Specifically, bluer markers occurred earlier in the trial, whereas yellower markers occurred later. Diamond markers show all fixations that occurred within ±1s around speech onset (temporal filter). The magenta circle indicates the mouse-click location registered by the participant at the end of the trial. The red cross shows the coordinate-wise median location of all diamonds. The red dashed circle marks a region of 2 dva radius centered on the median. This is our spatial filter. Any diamonds that were within this red dashed circle were sent to the VLM, and the rest were discarded. Except for those trials where the spatial filtering process resulted in zero remaining fixations, in which case all temporally filtered fixations, or diamonds, were sent to the model.

This visualization enables us to easily grasp how eye gaze evolves through time as a function of task instructions. Several features of the gaze data are immediately apparent. First, subjects take different amounts of time for different trials; however, the duration of speech remains more or less the same across trials. Second, the longer the time taken in a trial, the more fixations are made. Third, although fixations closer in time usually cluster together, there are also fixations made later on the same objects visited previously. Finally, and most importantly, fixations made around the time of speech (or question asked) generally indicate the location of the object eventually reported by the subject with a mouse click. In other words, people look at the object they are asking about around the time they utter the question.

Although our temporal and spatial filtering algorithm did well for the most part, achieving successful disambiguation in ~88\% of images overlaid with eye gaze data, there were a few cases where it failed (see Section~\ref{sec:appendix3}). As described in the main text, some of the original failures were caused by our spatial filtering process, resulting in zero remaining fixations. For these cases, all fixations that remained after temporal filtering were sent to the VLM. This led to a 1.5\% improvement in model performance, measured as the accuracy of model response. 

Next, we show some cases of failures in Figure~\ref{fig:figA1b}, including cases where there were zero remaining fixations after employing a median-based spatial filter (right top and right bottom panels). There are several causes of these failures: first, the object in question may have been fixated too few times leading to a shift in the median towards the wrong referent (top left panel in Figure~\ref{fig:figA1b}), or there might have been too few fixations close to speech onset on the correct object (bottom left), or too many fixations overall leading to an uninformative median (top and bottom right), or the model might have been incorrect even with the right eye gaze data (detection error, as described in Section~\ref{sec:appendix3}). 

Finally, we present examples of the 2\% cases where median filtering resulted in zero fixations, and retaining all fixations that occurred within ±1s of speech onset (essentially removing the spatial filter) yielded a more accurate VLM response (Figure~\ref{fig:figA1c}).

\subsection{Speech Behavior and Fixation Descriptive Statistics}\label{sec:appendix2}

We present more insights into the speech (questioning) behavior of the subjects in the study, and also discuss the descriptive statistics of the fixations collected in our data.

\begin{figure*}[t]
    \centering
    \includegraphics[width=0.9\textwidth]{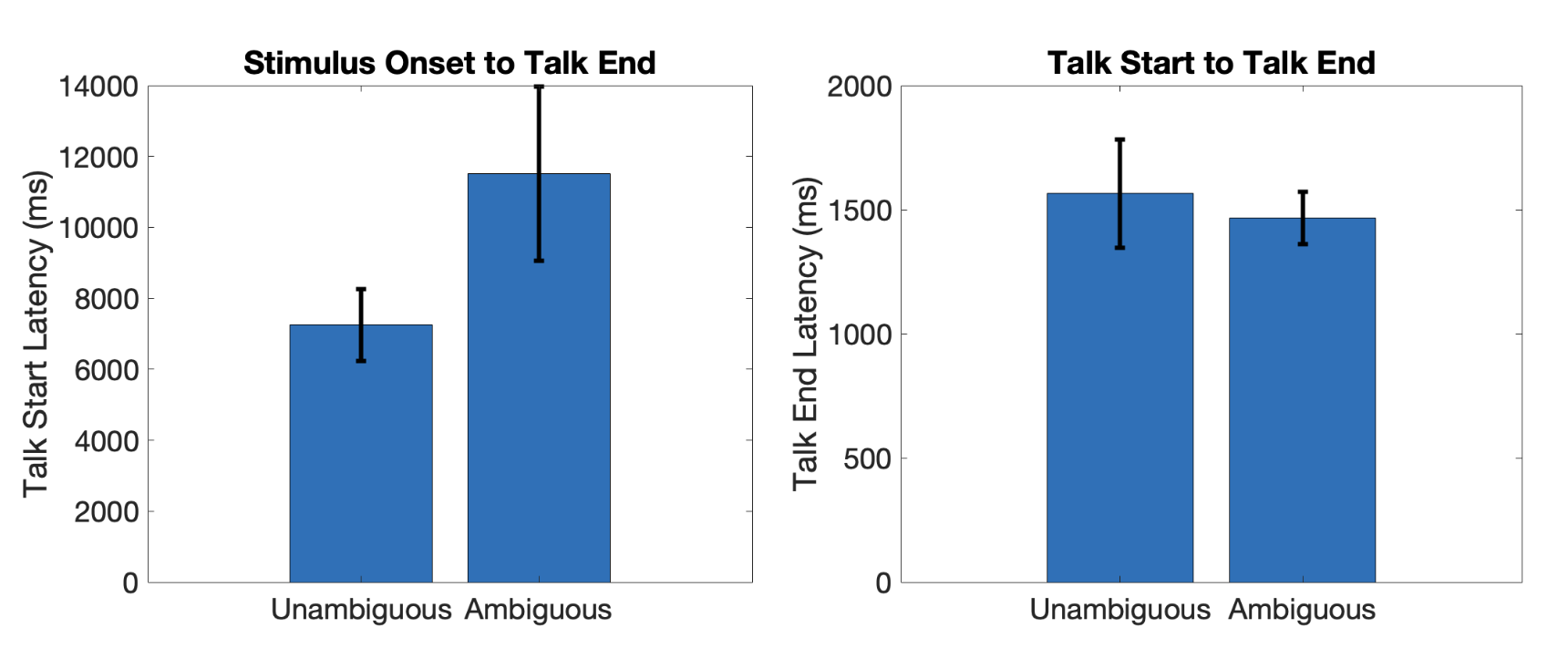}
    \caption{\textbf{Participants' speech (questioning) behavior.} The left panel shows the mean time taken by participants to formulate and ask a question from the time of stimulus onset. The right panel shows the mean duration between speech onset and end, as a function of experiment condition. The error bars indicate the standard error of the mean.}
    \label{fig:figA2a}
\end{figure*}

First, Figure~\ref{fig:figA2a} shows the average time taken by participants to ask a question once the trial starts (left) and the average time between speech onset and end (in other words, the mean duration of speech during asking the question) (right). Although participants took longer to ask a question in the Ambiguous condition (M=11.51s, SE=2.45s) than in the Unambiguous condition (M=7.25s, SE=1.02s) on average, this difference was not significant ($t(9)=-1.87, p=0.09$). The mean duration of questions asked was also not significantly different between Ambiguous (M=1.47 s, SE=0.11 s) and Unambiguous (M=1.56 s, SE=0.22 s) conditions ($t(9)=.46, p=0.66$).

\begin{figure*}[t]
    \centering  \includegraphics[width=0.9\textwidth]{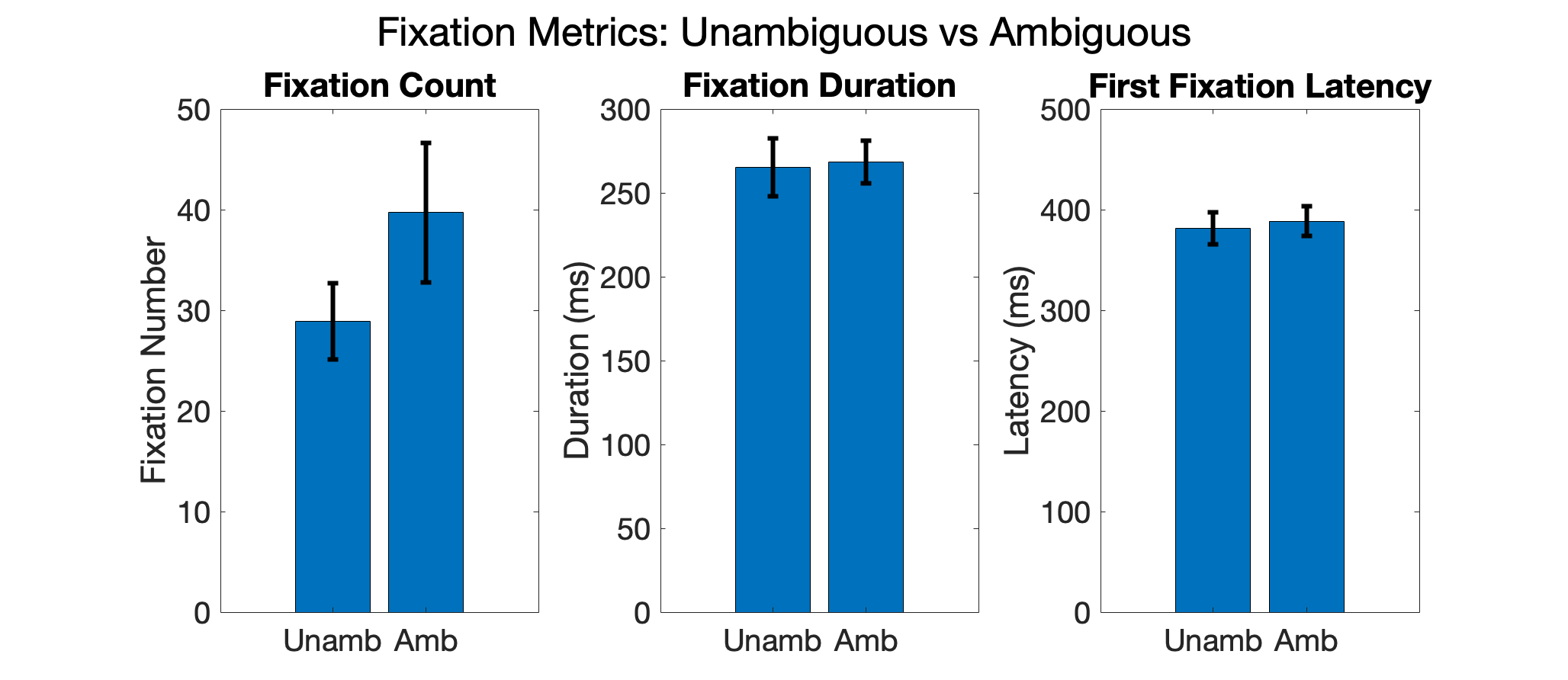}
    \caption{\textbf{Fixation metrics.} From left to right, the panels show average fixation count, fixation duration, and first fixation latency for Unambiguous and Ambiguous conditions. The error bars indicate the standard error of the mean.}
    \label{fig:figA2b} 
\end{figure*}

Next, Figure~\ref{fig:figA2b} shows some descriptive statistics for the fixations observed in our study. The leftmost panel shows the average fixation number for our two conditions. This aligns with the longer time between the stimulus onset and the start of questioning observed for ambiguous trials: a longer time allows for more fixations. Although the Ambiguous condition has a higher mean number of fixations (M=39.71, SE=6.95) than the Unambiguous condition (M=28.91, SE=3.79), this difference is not significant ($t(9)=-1.71, p=0.121$). The middle panel shows the mean duration of a given fixation on unambiguous and ambiguous trials. On average, fixation durations were highly similar between Ambiguous (M=268.31ms, SE=12.74ms) and Unambiguous conditions (M=265.06ms, SE=17.30ms) ($t(9)=-0.43, p=0.675$). Finally, the rightmost panel shows the mean latency of the first fixation after stimulus onset. As observed for fixation duration, the average first fixation latency was very similar between Ambiguous (M=388.62ms, SE=14.47ms) and Unambiguous conditions (M=381.32ms, SE=15.96ms) ($t(9)=-0.51, p=0.621$).

\subsection{Taxonomy of VLM Response Errors: Unsuccessful Trials}\label{sec:appendix3}

We analyze the "unsuccessful" trials to quantify why some trials fail in terms of the various types of errors observed in the different stages of the question-answering process. To do this, we manually categorized each of the 1500 VLM responses (50 stimulus images X 10 subjects X 3 categories - Image+LOI, Image+Gaze, Image Only) into four types of errors. The types of errors, listed roughly in the temporal order of their potential occurrence, are elaborated below along with some examples (see Figure~\ref{fig:figA3a}):

\begin{figure*}[t]
    \centering
    \includegraphics[width=0.9\textwidth]{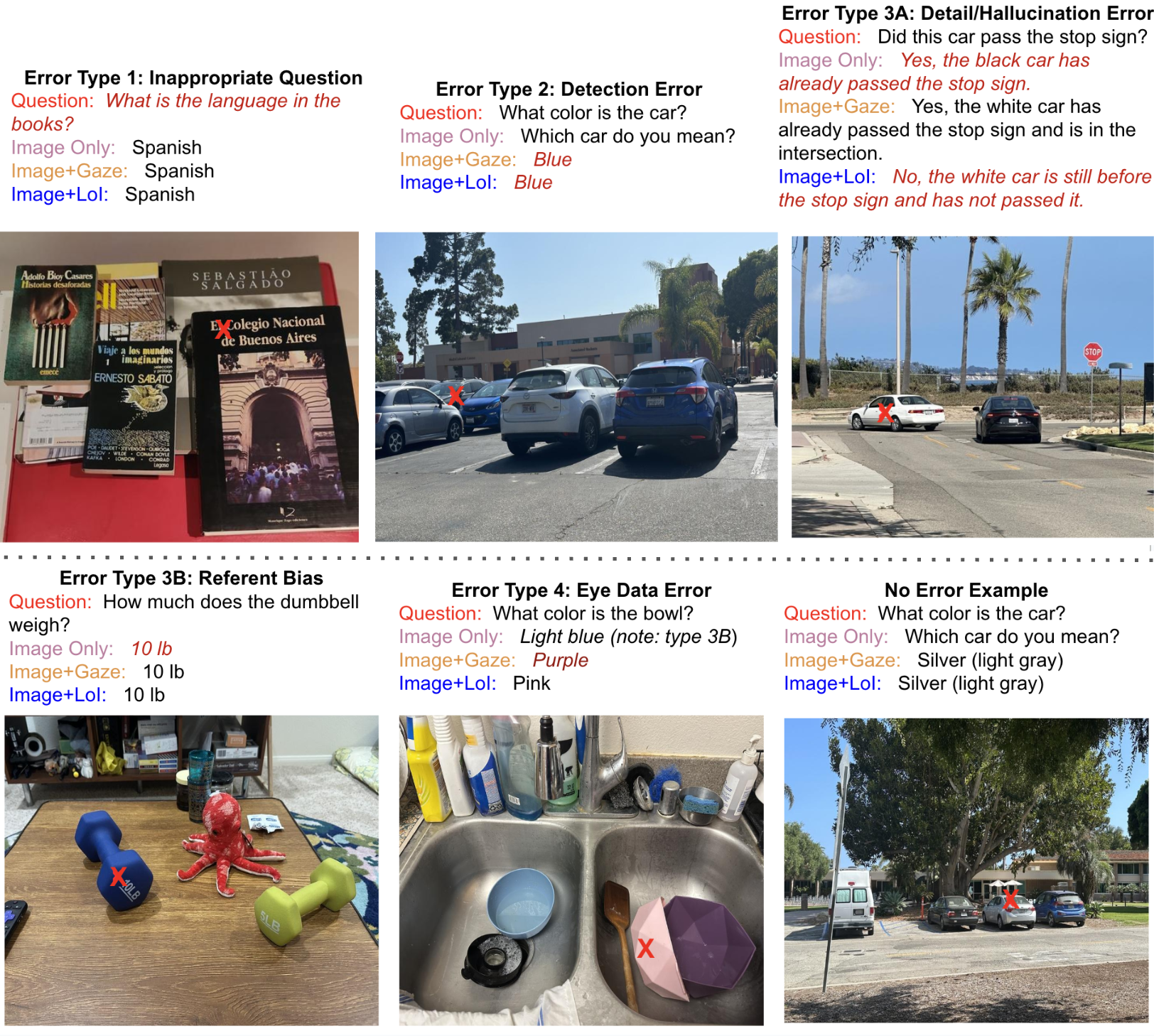}
    \caption{\textbf{Types of errors.} Example trials showing Inappropriate Question, Detection Error, Detail/Hallucination Error, Referent Bias, and Eye Data Error (discussed in Section~\ref{sec:appendix3} in detail). The bottom right panel shows an example of a trial where none of the types of errors occurred. The red cross in each image indicates the mouse click location, depicting the object asked about in a given trial.}
    \label{fig:figA3a}
\end{figure*}

\begin{itemize}
    \item Error Type 1. Inappropriate Question. In this case, the failure happened because the participant asked a question that was not suitable for the purposes of testing whether eye gaze data can help disambiguate visual information in the presence of ambiguity. For example, asking about the color of a bottle when there is only one bottle in the scene for a trial in the Ambiguous condition. Around 5.2\% of the trials had this kind of error.
    \item Error Type 2. Detection Error. In this case, an appropriate question was asked by the participant, yet the model failed to answer the question accurately when eye gaze data or mouse click locations for ambiguous trials were sent to the VLM, or for unambiguous trials with or without eye gaze data and mouse click locations. Around 5.1\% of the trials had this kind of error.
    \item Error Type 3. At this stage, the participant had asked an appropriate question, but one of the following things happened, depending on whether eye gaze data or mouse-click location was sent to the model along with the image: 
    \begin{itemize}
        \item Hallucination/Detail Error (Image+LOI or Image+Gaze). In this case, the VLM correctly detected the object category being asked about, yet responded with an incorrect detail and/or hallucinated. Approximately 2\% of all 'Image+LOI' and 'Image+Gaze' trials had this type of error.     
        \item Referent Bias (ambiguous Image Only trials). In this case, the VLM assumes one of the multiple referents was being asked about and answers with the details of that object. This reflects the model’s tendency to select a particular instance as the intended referent when multiple same-category instances are present. To be clear, when no eye gaze data is sent to the VLM, an accurate response of the VLM could either be a follow-up clarification question or a listing of the details of all the instances of a category (see bottom right "No Error" example in Figure~\ref{fig:figA3a}). Around 57\% of all 'Image Only' trials had this kind of bias.       
    \end{itemize}
    \item Error Type 4. Eye Data Error. The final form of failure is observed when the VLM response is accurate for the Image+LOI subtype but inaccurate for the Image+Gaze subtype for the same trial (image-question pair). This implies that the portion of the gaze data sent to the VLM for that trial might not have accurately captured the object being asked about. 11.6\% of all 'Image+Gaze' trials had this kind of error. Importantly, only ambiguous trials could have this kind of error because eye gaze data would be uninformative for unambiguous trials.
\end{itemize}

\begin{figure*}[t]
    \centering
    \includegraphics[width=1\textwidth]{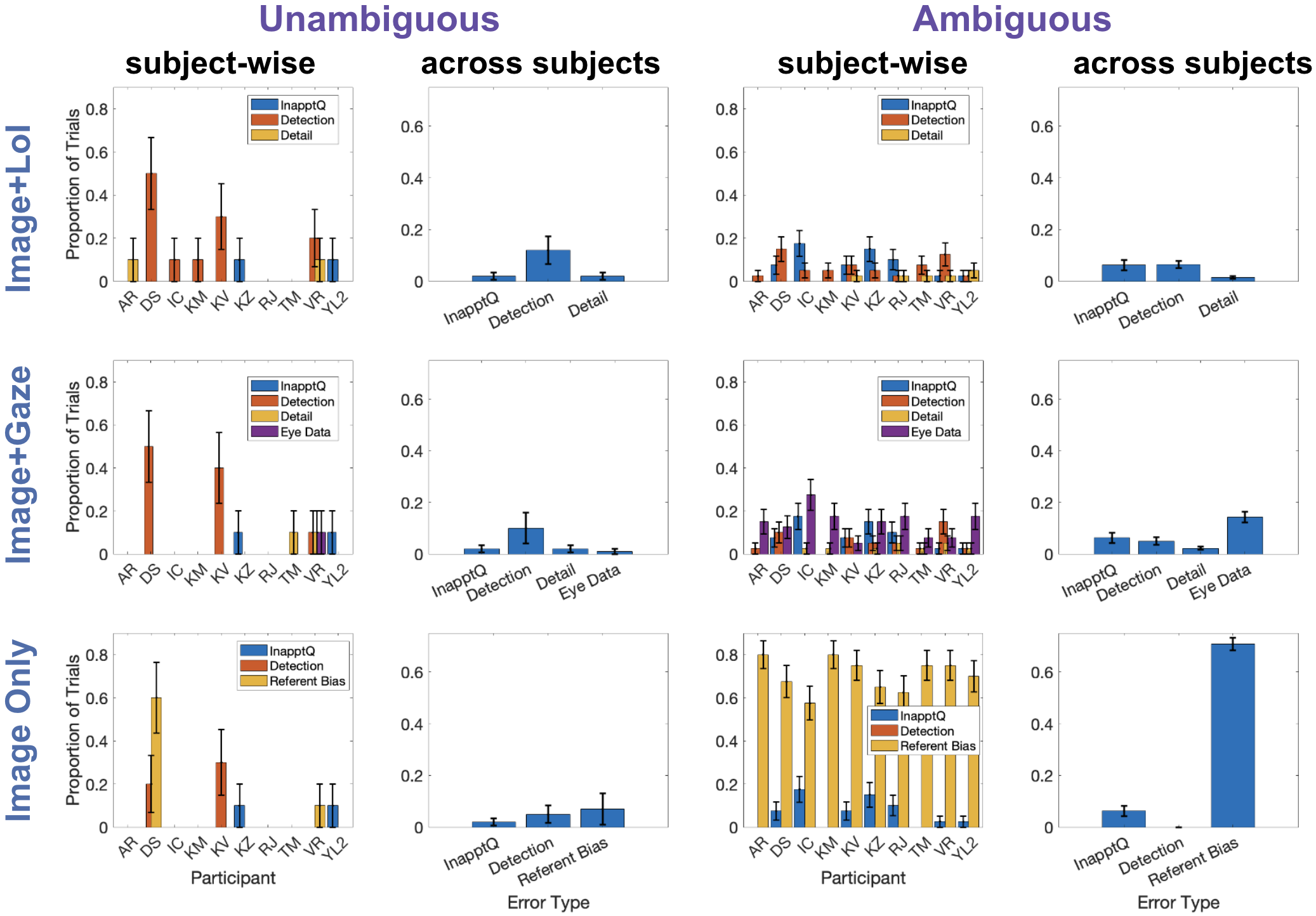}
    \caption{\textbf{Quantification of error types.} Proportion of trials that suffered from the error types discussed above and exemplified in Figure~\ref{fig:figA3a}. The three rows show data from Image+LOI, Image+Gaze, and Image Only input categories from top to bottom, respectively. The four columns show data from Unambiguous and Ambiguous conditions, each separated into subject-wise and across-subject measures. InapptQ - Error Type 1 (Inappropriate Question), Detection - Error Type 2, Detail - Error Type 3A (Detail/Hallucination Error), Referent Bias - Error Type 3B, Eye Data - Error Type 4.}
    \label{fig:figA3b}
\end{figure*}

Figure~\ref{fig:figA3b} shows the proportion of trials within a specific condition (Ambiguous or Unambiguous) and a specific VLM input type (Image+LOI, Image+Gaze, and Image Only) with the different error types as discussed above, both as subject-wise and across subjects measures. It is worthwhile to note that most of the errors of type 1 (inappropriate questions) were contributed by only a few of the subjects, such as IC and KZ. Secondly, for the Unambiguous condition, Detection Error forms the major contributor to unsuccessful trials, when mouse click location or eye gaze data is sent to the VLM. For the Ambiguous condition, a major portion of the unsuccessful trials is caused by error type 4 (Eye Data Error), which is when the VLM responds accurately when prompted with Image+LoI but not with Image+Gaze. Finally, for ambiguous trials, when only the image without any mouse clicks or eye gaze data is sent to the model, a major portion (70\%) of the responses contain type 3B errors (Assumption Error). This is in line with the findings of \citet{testoni2024racquet} who show that LLMs have a bias in responding with stereotypical, and sometimes risky, assumptions when prompted with inherently ambiguous questions.

\subsection{List of System Prompts}\label{sec:appendix0}
\textbf{VQA prompt for Image+Gaze/LOI input: }\textit{"You are an expert visual interpreter specialized in identifying and describing features of specific objects in images by considering the eye movement data. The user will provide an original image alongside the same image with eye movement data, where fixation points indicated by white X signs. Alongside these two images, the user will ask a specific question regarding an object's features, such as its color, size, shape, or spatial location relative to other objects present in the image.
Your task is to carefully analyze the image, identify the specific object in question considering the eye fixation points, and answer the user's question precisely and concisely. Your response should only describe the referred object closest to the fixation points indicated by white X signs. You should provide clear, factual answer without any extra output. Do not speculate or include details unrelated to the indicated object or question. Do not mention the eye movement data in your response.
Your response should be very short, concise, but accurate."}

\textbf{VQA prompt for Image-Only input: }\textit{"You are an expert visual interpreter specialized in identifying and describing features of specific objects in images. The user will provide an image, and alongside that image, the user will ask a specific question regarding an object's features, such as its color, size, shape, or spatial location relative to other objects present in the image.
Your task is to carefully analyze the image, identify the specific object in question, and answer the user's question precisely and concisely. You should provide clear, factual answer without any extra output. Do not speculate or include details unrelated to the indicated object or question.
Your response should be very short, concise, but accurate."}

\textbf{VQA prompt for generating "wrong answer": }\textit{"You are an expert visual interpreter specialized in identifying features of specific objects in images by considering the eye movement data. The user will provide an original image alongside the same image with eye movement data, where fixation points indicated by white X signs. Alongside these two images, the user will ask a specific question regarding an object's features, such as its color, size, shape, or spatial location relative to other objects present in the image.
Your task is to carefully analyze the image, identify the specific object in question considering the eye fixation points, and answer the user's question concisely but wrongly. Your response should only be the wrong answer about the referred object closest to the fixation points indicated by white X signs. You should provide clear, wrong answer without any extra output. Do not speculate or include details unrelated to the indicated object or question. Do not mention the eye movement data in your response.
Your response should be very short, concise, but wrong."}

\textbf{VLM accuracy evaluation prompt: }\textit{"You are an expert visual interpreter specialized in identifying the correctness of an answer to a question about an object in the image. The user will provide an image alongside the same image with the referent object in question indicated by a white X sign. Alongside these two images, the user will provide the specific question and answer.
Your task is to analyze the images and decide whether the answer about the referent object is correct or not. Your response should only include one word: "correct" or "incorrect". Do not include any other words or details in your response."}

\subsection{Participant Task Instructions}
Participants received the following verbal instructions at the start of the experiment: \textit{``In this study, you will view a series of images on the screen. For each image, your task is to formulate a question about a specific object in the scene and ask it out loud. In some blocks, you will be asked to formulate questions that could refer to multiple objects (ambiguous), while in other blocks, your questions should clearly refer to only one object (unambiguous). After you ask your question, you will hear a response from an AI assistant. Finally, you will click on the object you were asking about. Please keep your head still and positioned on the chinrest throughout the experiment. You may take breaks between blocks if needed.''} Participants then completed 7 practice trials with feedback before beginning the main experiment.


\subsection{Related Work}
\textbf{Vision-Language Models and Visual Question Answering.}
VLMs have progressed from early two-stream transformers (VisualBERT \citep{li2019visualbert}, LXMERT \citep{tan2019lxmert}) and contrastive pretraining (CLIP; \citep{radford2021clip}) to instruction-tuned systems (Flamingo \citep{alayrac2022flamingo}, BLIP-2 \citep{li2023blip2}, LLaVA \citep{liu2023llava}) and recent open models that markedly advance VQA and multimodal reasoning: LLaVA-OneVision \citep{li2024llavaonevision}, Qwen2-VL and Qwen2.5-VL \citep{wang2024qwen2vl,bai2025qwen2_5_vl}, InternVL~2.5 \citep{chen2024internvl25}, IDEFICS2 \citep{laurencon2024idefics2}, and large-scale pretraining analyses such as MM1 \citep{mckinzie2024mm1}.
Yet grounding issues persist, including failures owing to visual shortcomings in MLLMs \citep{tong2024eyeswideshut} or bypassing visual grounding in shortcut learning \citep{reich2024role}. When questions are underspecified, VQA performance may be helped by prompt-level fixes (e.g., visually grounded rephrasing) \citep{prasad2024rephraseaugmentreasonvisual}.
Complementing these trends, emerging work integrate human gaze cues within images to clarify referents by introducing new model architectures \citep{inadumi2024gazevqa}. In contrast, our approach leverages the viewer’s natural gaze at inference, without architectural changes or retraining, to guide the model toward the intended referent and reduce ambiguity in open-ended VQA.

\noindent\textbf{Eye Movements and Attention.}
Foundational research links eye movements to covert attention \citep{posner1980orienting,hoffman1995role,deubel1996saccade,li2021dissociating}. In natural tasks, gaze anticipates actions and speech, revealing time-resolved planning during sentence production \citep{coco2012scan}, and scene and event description \citep{hayhoe2012predictive,griffin2000eyes}. Classic studies have demonstrated that what observers look at depends strongly on visual and linguistic context and goals \citep{yarbus1967eye,tanenhaus1995integration}. While this temporal coupling is well established, it is rarely \emph{operationalized} to select referents at the precise moment of question formulation in VQA. This work aligns a short window around speech onset with fixations to harvest the most informative disambiguation cue.

\textbf{Eye-Tracking in Vision-Language and Computer Vision.}
Eye movements have a rich history of informing models of perception and learning. \citet{yarbus1967eye} pioneered the use of eye movements to understand visual perception, showing that fixation patterns vary with task demands. In the context of language processing, \citet{tanenhaus1995integration} demonstrated that eye movements during spoken language comprehension reflect real-time semantic processing. Interactive systems have long explored gaze as an input modality: \citet{jacob1991use} introduced early gaze-based interaction paradigms, and \citet{majaranta2014eye} surveyed the evolution of eye-tracking interfaces. 
Building on these foundations, the computer vision community has increasingly recognized the value of gaze data: classical saliency models codify attention mechanisms inspired by human vision \citep{itti2001computational} (see \citet{borji2013state} for a review), and gaze has been used to directly guide concrete vision tasks. For example, \citet{shanmuga2015eye} used multi-viewer eye-tracking to guide video object segmentation, showing that gaze can directly support object localization and extraction. More recently, gaze has been integrated as a supervisory signal in learning-based systems. For instance, \citet{sugano2016seeing} utilized gaze supervision to enhance image captioning models. 

Within VQA, \citet{sood2021gazevqa} introduced VQA-MHUG, a gaze-annotated dataset designed to study multimodal neural attention, and \citet{ilaslan2023gazevqa} presented GazeVQA, a video QA dataset capturing multiview gaze in task-oriented collaboration scenarios. Building on these directions, our work creates a real-time system that naturally integrates speech, gaze, and vision to resolve ambiguous questions in open-ended VQA. \citet{inadumi2024gazevqa} used gaze-target estimations to improve VQA performance, and \citet{karessli2017gazeembeddingszeroshotimage} used human gaze embeddings as auxiliary information for zero-shot image classification. However, most prior work leverages gaze during \emph{training} or as estimated gaze target annotations, and model-internal attention is not a reliable substitute for human referential intent. Our work creates a real-time system that naturally integrates human fixations, speech, and visual inputs directly \emph{at inference} to bias VLM reasoning without retraining or architectural changes.

\textbf{Multimodal Ambiguity Resolution.}
Ambiguity resolution has been studied across modalities. In NLP, \citet{sukthanker2018anaphora} surveyed approaches to anaphora resolution (identifying what a referring expression refers to in context). In computer vision, \citet{mao2016generation} addressed ambiguous referring expressions by generating unambiguous descriptions, and \citet{kazemzadeh2014referitgame} created a game to collect referring expressions for objects in images. At the VQA intersection, \citet{zhang2016yin} introduced a balanced dataset for binary ("yes/no") VQA on abstract scenes rather than ambiguous questions, while more recently \citet{testoni2024racquet} highlighted issues with overconfidence and bias in visual LLMs under ambiguity. Against this backdrop, our approach uniquely combines real-time gaze tracking with open-ended VQA to address ambiguity in a naturalistic setting.

\subsection{Ethics statement}
We conducted this research in accordance with the ACL Code of Ethics and applicable regulations. This study involving human participants was reviewed and approved by the human subject research protocols (to maintain anonymity, details will be provided upon acceptance). All participants provided informed consent prior to participation and could withdraw at any time without penalty. The consent form explained that: eye movements and voice recordings would be collected during the session, data would be pseudonymized and used for research, participation was voluntary and could be discontinued at any time without penalty, and anonymized data may be shared publicly for research reproducibility, excluding any personally identifiable information such as raw audio recordings. 

Recruitment occurred via our university subject pool; participants received course credit. 
We collected monocular eye-tracking data at 1000\,Hz, participants' voices, and mouse click locations for research purposes only. Personally identifying information was not collected; data were pseudonymized before sharing in compliance with institutional and legal requirements. The public release data will include only fixation coordinates and the transcribed text format of the audio to reduce re-identification risks; raw audio or any Personally Identifiable Information (PII) will not be released.
Potential risks to participants were minimal and limited to transient discomfort/fatigue; these were mitigated by calibration breaks, screen distance guidelines, and data minimization. The participant sample (primarily students) may limit demographic diversity; we report results with this limitation in mind and caution against over-generalization.

\subsection{Reproducibility Statement}
To preserve anonymity, we currently release only the practice trials of our VQA benchmark (containing only MS COCO images), as well as semantic embeddings of model responses, and an example gaze-visualization code.
The full stimulus image set, VLM responses, real-time interactive VQA demo, and the complete code of the project will be released upon acceptance.
Our anonymized supplementary material can be accessed from:
\url{https://drive.google.com/file/d/1Ly5WGakEy_1MnjqW2WPjGQ-DVMAg05TS/view?usp=share_link}

\subsection{LLM Use Disclosure}
We used LLMs to correct grammatical mistakes, polish writing, clean software code, and search related works.

\end{document}